\documentclass[journal]{IEEEtran}

\usepackage{cite}

\ifCLASSINFOpdf
\else
\fi

\usepackage[cmex10]{amsmath}
\usepackage{algorithm}
\usepackage{algorithmic}

\usepackage{array}

\ifCLASSOPTIONcompsoc
  \usepackage[caption=false,font=normalsize,labelfont=sf,textfont=sf]{subfig}
\else
  \usepackage[caption=false,font=footnotesize]{subfig}
\fi
\usepackage{fixltx2e}
\usepackage{url}

\usepackage{times}
\usepackage{epsfig}
\usepackage{graphicx}
\usepackage{xcolor}
\usepackage{amssymb}
\usepackage{makecell,rotating}
\usepackage{algorithm}
\usepackage{algorithmic}
\usepackage{multirow}
\usepackage{booktabs}

\renewcommand{\(}{\left(}
\renewcommand{\)}{\right)}

\DeclareMathOperator*{\argmax}{arg\,max}
\DeclareMathOperator*{\minimize}{minimize}

\renewcommand{\b}{\mathbf{b}}
\newcommand{\x}{\mathbf{x}}
\newcommand{\y}{\mathbf{y}}
\newcommand{\p}{\mathbf{p}}
\renewcommand{\c}{\mathbf{c}}
\renewcommand{\i}{\mathbf{i}}
\renewcommand{\j}{\mathbf{j}}
\newcommand{\w}{\mathbf{w}}

\graphicspath{{fig/}{portrait_bio/}}

\begin{document}
%
\title{Unified Structured Learning for Simultaneous Human Pose Estimation and Garment Attribute Classification}
%
%
%

\author{Jie~Shen,
        Guangcan~Liu,~\IEEEmembership{Member,~IEEE},
        Jia~Chen,
        Yuqiang~Fang,
        Jianbin~Xie,~\IEEEmembership{Member,~IEEE},
        Yong~Yu,
        and~Shuicheng~Yan,~\IEEEmembership{Senior~Member,~IEEE}

\thanks{Jie Shen is with the Department of Computer Science and Engineering, Shanghai Jiao Tong University, Shanghai, 200240, China (e-mail: jieshen@apex.sjtu.edu.cn).
}

\thanks{Guangcan Liu is with the School of Information and Control Engineering, Nanjing University of Information Science and Technology, Nanjing, 210044, China. E-mail: gcliu@nuist.edu.cn}

\thanks{Jia Chen and Yong Yu are with the Department of Computer Science and Engineering, Shanghai Jiao Tong University, Shanghai, 200240, China.

\noindent
E-mail: chenjia@apex.sjtu.edu.cn; yyu@apex.sjtu.edu.cn}

\thanks{Yuqiang Fang is with the College of Mechatronic Engineering and Automation, National University of Defense Technology, 410073, China. 

\noindent
E-mail: fangyuqiang@nudt.edu.cn}

\thanks{Jianbin Xie is with the College of Electronic Science and Engineering, National University of Defense Technology, China. 

\noindent
E-mail: jbxie@126.com}

\thanks{Shuicheng Yan is with the Department of Electrical and Computer Engineering, National University of Singapore, Singapore. 

\noindent
E-mail: eleyans@nus.edu.sg}
}

%
%

\markboth{IEEE TRANSACTIONS ON IMAGE PROCESSING, VOL. XX, NO. XX, XX 2014}%
{Shell \MakeLowercase{\textit{et al.}}: Bare Demo of IEEEtran.cls for Journals}
%



\maketitle

\begin{abstract}
In this paper, we utilize structured learning to simultaneously address two intertwined problems: 1) human pose estimation (HPE) and 2) garment attribute classification (GAC), which are valuable for a variety of computer vision and multimedia applications. Unlike previous works that usually handle the two problems separately, our approach aims to produce an optimal joint estimation for both HPE and GAC via a unified inference procedure. To this end, we adopt a preprocessing step to detect potential human parts from each image (i.e., a set of ``candidates'') that allows us to have a manageable input space. In this way, the simultaneous inference of HPE and GAC is converted to a structured learning problem, where the inputs are the collections of candidate ensembles, the outputs are the joint labels of human parts and garment attributes, and the joint feature representation involves various cues such as pose-specific features, garment-specific features, and cross-task features that encode correlations between human parts and garment attributes. Furthermore, we explore the ``strong edge'' evidence around the potential human parts so as to derive more powerful representations for oriented human parts. Such evidences can be seamlessly integrated into our structured learning model as a kind of energy function, and the learning process could be performed by standard structured Support Vector Machines (SVM) algorithm. However, the joint structure of the two problems is a cyclic graph, which hinders efficient inference. To resolve this issue, we compute instead approximate optima by using an iterative procedure, where in each iteration the variables of one problem are fixed. In this way, satisfactory solutions can be efficiently computed by dynamic programming. Experimental results on two benchmark datasets show the state-of-the-art performance of our approach.
\end{abstract}


%
\IEEEpeerreviewmaketitle

\section{Introduction}\label{sec:intro}
%
%
%
%

\IEEEPARstart{H}{uman-oriented} technologies play important roles in many computer vision and multimedia applications that
require interactions between persons and electronic devices. The
significance of human-oriented technologies naturally drives the research community to extensively investigate human-related topics, such as face
recognition~\cite{sdm}, human tracking~\cite{p-track}, pose estimation~\cite{deva11}, clothing technology~\cite{clotheccv}, etc. In
this work, we are interested in two of them: human pose estimation (HPE) and
clothing technology (CT). Both problems have been studied extensively, and a review of previous
work is presented in the following section.

\begin{figure}[tbp]
\centering
\includegraphics[width=0.99\linewidth]{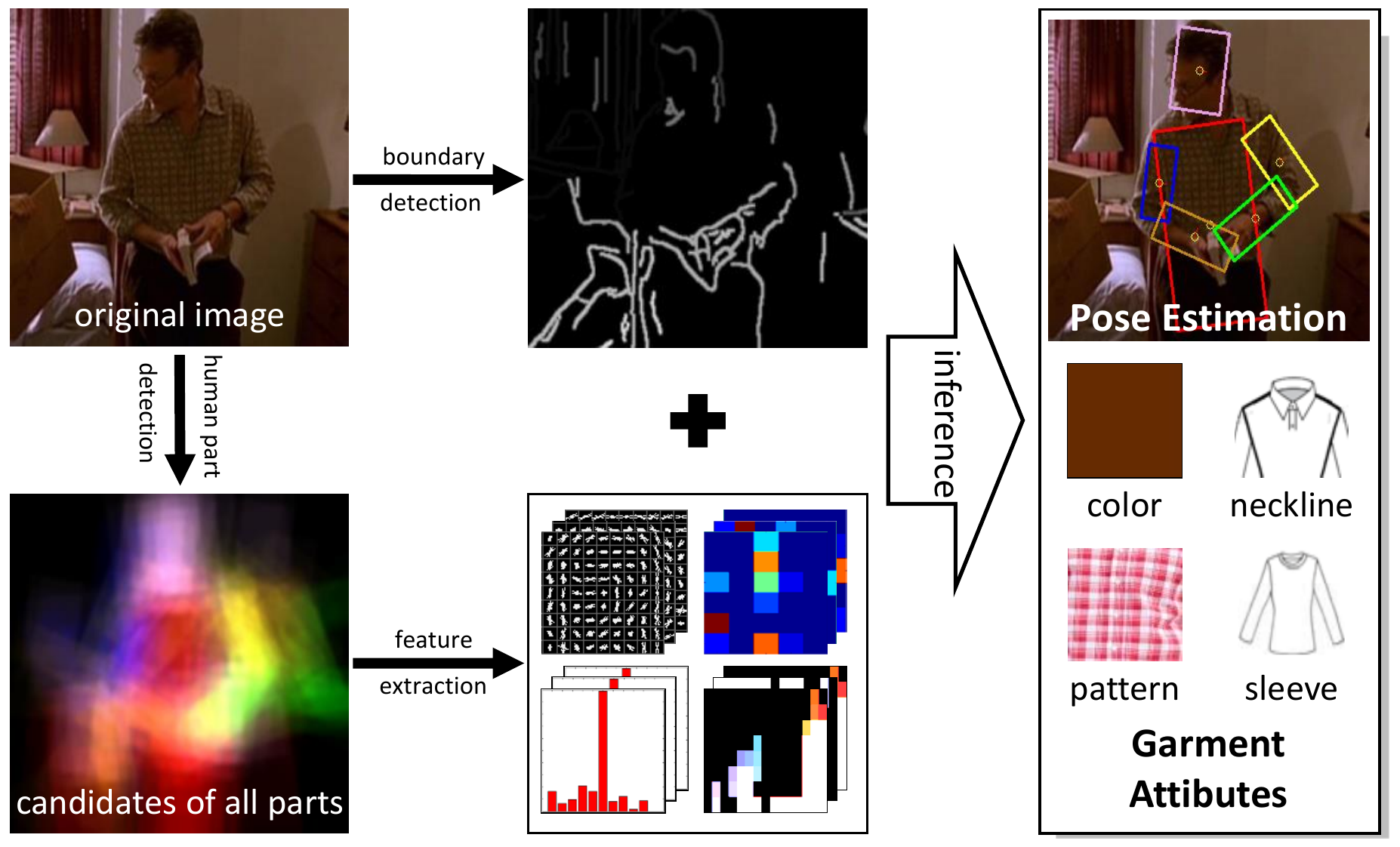}
\caption{
\textbf{Overview of the proposed approach.} For a given image, first we detect candidates
that have potentials to be valid human parts using the HPE algorithm proposed
in~\cite{deva11}. This preprocessing step simplifies the representation of the image
and converts the joint inference of HPE and GAC into a structured learning problem.
Second, we design the joint feature representation for structured learning using
various visual cues. Meanwhile, potential strong edges in the image are detected by utilizing
a well-established algorithm presented in~\cite{gpb}.
 Finally, we use the prediction function learned from structured
SVM to produce the joint labels of human parts and garment attributes.}
\label{fig:frame}
\end{figure}

\subsection{Previous Work}\label{subsec:related}
\subsubsection{Human Pose Estimation}
The literature about HPE can trace back to 40 years ago. Fischler and
Elschlager~\cite{ps1} proposed to represent the articulated human pose as a
collection of rigid body parts. This classical model, called \emph{pictorial
structure} (PS) in \cite{ps2}, provides a straightforward representation for articulated objects and owns a tree
structure that can facilitate efficient inference. Hence, PS is still adopted as a basic tool by recently
established approaches, e.g.,~\cite{cvpr09, bmvc09, dpm, fer08, nips06, songchun}.
These recent works mainly pursue better feature description, more efficient computation, more complex human structures and more
effective contextual information.

Feature description is one of the key elements for various vision tasks and so HPE~\cite{xm}. In~\cite{nips06}, an iterative parsing paradigm
was introduced to obtain an increasingly finer feature scheme for describing human parts. Other works, e.g.,~\cite{cvpr09, dpm}, employed shape-based
feature descriptors such as Shape Context~\cite{shape-context} and Histogram of Oriented Gradients (HOG)~\cite{hog}, which proved to be more effective than color-based features. In~\cite{bmvc09}, Eichner and Ferrari considered the appearance consistence between
connected/symmetric limbs and developed a better appearance model.

The majority of early works on HPE emphasized on detection performance, i.e., a more
effective inference schema, and the focus of later works was placed on the efficiency of inference.
Typically, the search space of human pose is the main bottleneck for improving the efficiency, because one must search over the location and orientation space for each human part, as well as over the image pyramid.
To reduce the search space of human poses, Ferrari et al.~\cite{fer08} utilized a generic upper-body detector
and the GrabCut~\cite{grabcut} algorithm,
targeting at a reduced search space. In~\cite{cvpr09, cvpr10, eccv10}, the tree structure is used to make the
inference procedure more efficient, which
can also well handle the spatial association between different human parts.
One advantage of the tree structure is the ability to allow a fast computation via a dynamic programming~\cite{ps2}.
Furthermore, deformable cost computation between connected human parts can be accelerated by performing
a distance transform if the pair-wise cost suffices some specific conditions~\cite{ps2}.

Although the tree-based PS models can draw a general representation for human body,
it does not explore the implicit connections between rigid parts that do not share joints~\cite{bb}.
Therefore, graph-based structures are further proposed and explored by recent works. Such models are hard to be learnt
exactly because of their high computational cost over the graph structure. Usually, Markov Random Filed (MRF) is used to achieve an approximate inference. By taking a branch and bound step in \cite{bb},
inference on a graph is nearly as efficient as the tree models.
Recently, Yang and Ramanan~\cite{deva11} proposed a method that represents each human part by
a mixture of templates. Unlike the previous limb models with articulated
orientations, their templates are non-oriented and can well capture near-vertical
and near-horizontal limbs. By tuning the part type and location, their model can handle the in-plane
rotation and foreshortening.

Another drawback of the original PS model is that the contextual information is not explicitly considered.
In the work of Sapp et al.~\cite{eccv10}, various visual cues (e.g., boundary and segmentation) were employed in their coarse-to-fine model.
In~\cite{songchun}, Rothrock et al. incorporated background context into the PS model. Their model encourages a high contrast of a part region from its surroundings. Experimental results in their paper demonstrated the effectiveness of such contextual information.

\subsubsection{{{Clothing Technology}}}
The clothing technologies, mainly including clothing
segmentation~\cite{clothseg1,clothseg2,cloth12}, clothing
recommendation~\cite{clothrec} and garment attribute
classification~\cite{clotheccv}, play an important role in many multimedia systems
such as clothing search engines~\cite{clothsearch}, online
shopping~\cite{clothliu} and human recognition~\cite{cloth08}.

The work of Chen et al.~\cite{cloth06} is one of the representative works in the related
literature, which introduced an And-Or graph to generate a large set of
composite clothing components for further recognition.
In \cite{clothliu}, Liu et al. proposed a cross-scenario clothing retrieval
system which can search similar garments from the online shop by using 
a person's daily life photo as input. In their work, a well trained human part detector was employed for part
alignment and an offline transfer learning scheme was introduced to handle the
discrepancy between images from daily life and online shops.
Recently, they proposed a practical system called ``magic closet'' in
\cite{clothrec} which can automatically recommend garments according to
occasions. They utilized the latent SVM algorithm and modeled the clothing
attributes as latent variables to provide mid-level features, rather than
directly bridging the raw image features to the occasions.

Recent progress in clothing techniques has witnessed the benefit of HPE due to
the strong relations between human parts and garment attributes. In other words, it is a
popular way to perform clothing study based on the results of human part detection
\cite{clotheccv, clothliu}.
Moreover, Yamaguchi et al.~\cite{cloth12} and Chen et al.~\cite{clotheccv} used
well trained human part detectors to produce a large number of garment types,
aiming to achieve precise clothing recognition.
Bourdev et al.~\cite{clothiccv} trained an SVM classifier over each human part
with the purpose of indicating whether or not a human part has specific garment
attributes.

\subsubsection{HPE and CT}
There are few works addressing the
interrelations between HPE and CT. Yamaguchi et al.~\cite{cloth12}
tried to refine both HPE and clothing parsing by using a three-stage scheme: first, they
obtained some initial results of HPE; second, they used those initial HPE
results as the basis to gain more reliable clothing segmentation; finally, the
produced segmentation results were used to further refine HPE. However, since
the quality of clothing segmentation largely depends on the success of HPE and vice
versa, such a separate modeling approach may fail to capture the correlations between the
two tasks and cannot achieve significant improvements over the competing
baseline, as can be seen from the reports in~\cite{cloth12}.
In the recent work of Ladicky et al.~\cite{ladicky}, they combined the part-based approach
of pose estimation and pixel-based approach of image labeling into a principle way so as
to inherit advantages of both. Inference for their model was performed with two steps: first,
they iteratively added the next best pose candidate by computing an energy function of their model;
second, they refined the final solution over the selected candidates of the first step.

\begin{figure*}[tbp]
\centering
\subfloat[]
{
\includegraphics[width=0.32\linewidth]{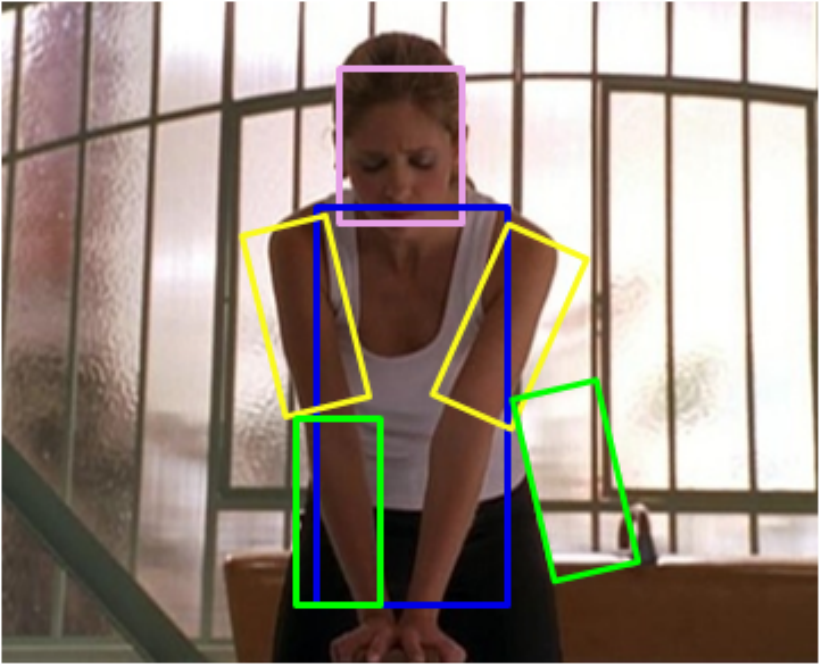}
}
\subfloat[]
{
\includegraphics[width=0.32\linewidth]{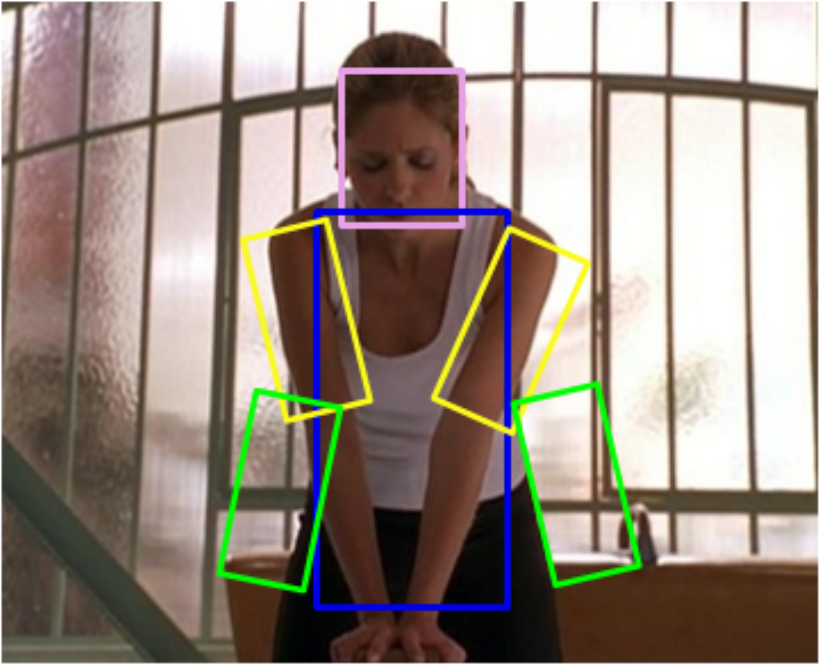}
}
\subfloat[]
{
\includegraphics[width=0.32\linewidth]{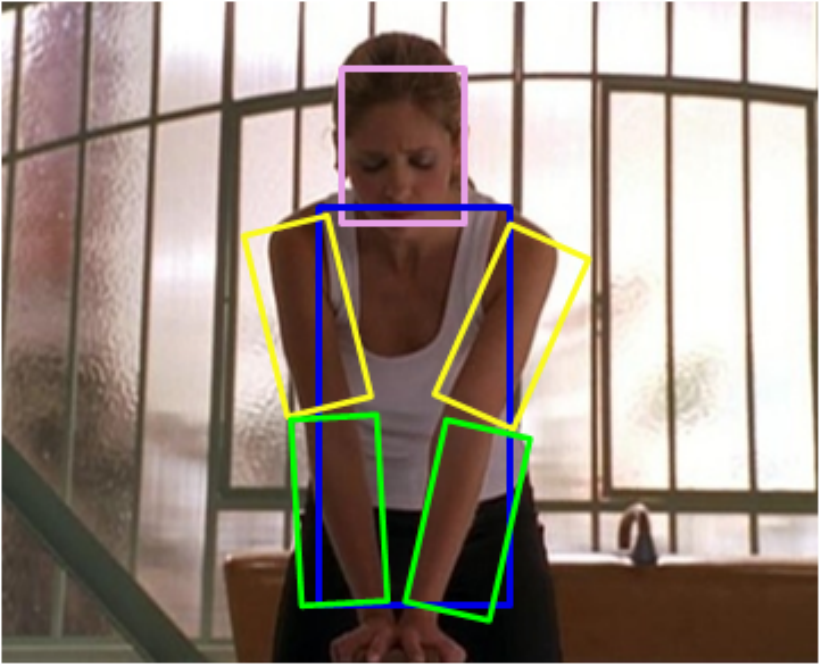}
}
\caption{Examples to illustrate that GAC can help HPE. (a) Result with incorrect
left-lower arm. (b) Result with both incorrect lower arms. (c) Result with all
correct arms. For (a), one can immediately assert that the HPE result can hardly
be correct because the appearances of right-lower arm and left-lower arm differ
greatly. Such prior knowledge about limb appearance
 was considered in
\cite{bmvc09}. However, it can't distinguish which human estimation is right from (b)
and (c) in this way, where the lower arms' appearances differ slightly.
Given that the garment attribute is known, e.g., sleeve type is sleeveless, one can easily
exclude (b) because the lower arms in (b) have few skin colors.
}
\label{fig:cp_for_hpe_example}
\end{figure*}
\subsection{Contributions of This Work}
It is indeed natural to anticipate that HPE and CT are intertwined problems and can
help each other. For example, in Figure~\ref{fig:cp_for_hpe_example}, depending on the
garment type, one can erase a large number of incorrect HPE candidates. However, existing approaches that individually use pose information to refine CT or use clothing knowledge to help HPE cannot fully capture the advantages
of modeling the correlations between the two tasks. In this paper, we therefore
propose to integrate HPE and garment attribute classification (GAC) into a unified framework, with the purpose of
making effective use of the possible correlations between human parts and
garment attributes.

Also, we aim to provide an effective way to jointly model multiple visual
cues, including the features specific for human parts (pose-specific
features) and for garment attributes (garment-specific
features), as well as the cross-task features that encode the correlations between
human parts and garment attributes.
For a more informative description for oriented human part, we
explore the ``strong edge'' evidence as an energy function so as to incorporate the
contextual information around a human part.
This motivation is based on the following observation: Since our representation for a human part is an oriented bounding box,
it generally holds that there exist parallel edges sharing a similar orientation with an
underlying correct part candidate, as illustrated in Figure~\ref{fig:sedge}.

To this end, we use the HPE algorithm presented in~\cite{deva11} to obtain from
each image a set of bounding boxes (called ``candidates'') that have potentials
to be correct human parts, resulting in a basic representation for each image --
one image is represented by one set of candidate ensembles. In this way, the joint
inference for HPE and GAC is converted to a \emph{structured
learning}~\cite{svm-struct} problem, where the input is the image represented
by a collection of candidate ensembles, the output is the joint labels of human
parts and garment attributes, and the joint feature representation involves the
aforementioned multiple visual cues.

Given a set of annotated training images,
the prediction function of structured learning is learnt by using the structured
Support Vector Machines (SVM) algorithm.
Inference for a new test samples can be performed efficiently by iteratively computing a local
optimum on a tree with the dynamic programming algorithm.
Experimental results on two benchmark datasets show the
state-of-the-art performance of our approach.

One may want to take HPE and GAC into the multi-task learning (MTL) framework~\cite{mtl}. We remark here that our problem cannot be solved via existing MTL methods since models of MTL always assume that the underlying tasks share the same feature space. In our case, however, this assumption is not valid as we address two different tasks: human pose estimation (a detection task) and garment attribute classification (a recognition task). Each of the two tasks has its own feature space that can not be shared with the other, i.e. pose-specific features and garment specific features (see Section~\ref{subsec:joint-feature}). Also, methods from domain adaption (DA)~\cite{da} cannot be applied as DA algorithms deal with the variations in some combinations of factors, including scene, object location and pose, view angle, resolution etc. Obviously, our problem is not under the setting of DA algorithms.

The rest of the paper is organized as follows. Section~\ref{sec:model} elaborates on
our approach for combined HPE and GAC, including feature design, parameter learning and inference algorithm. Section~\ref{sec:exp} presents the
experiments and results. Section~\ref{sec:con} concludes this paper and discusses
our future work.

\section{Joint Inference of HPE and GAC}\label{sec:model}

As Figure~\ref{fig:frame} shows, our approach contains three major procedures,
including a preprocessing step that detects candidates from each image, an
engineering step that forms a joint feature representation from various visual
cues, and an inference step that uses structured SVM learned from a set of training images. In the following, we shall detail each
step one by one.

\subsection{Preprocessing and Problem Formulation}
\label{subsec:notation}
We do not build our approach by directly utilizing the images represented by raw
pixels, and instead, we use the existing HPE method~\cite{deva11} to produce some
initial results as input to our approach.
More precisely, for each human part $i$ (e.g., head), we perform a non-maximum
suppression on the output of~\cite{deva11} and take the top $K_i$ (each $K_i=40$ in
our work) candidates (denoted by $\b_i$) from each image, where each candidate is a bounding box
$(x, y, \theta, s)$, with $(x, y)$, $\theta$ and $s$ denoting the center
coordinates, the angle and the size of the bounding box,
respectively~\footnote{The original output of~\cite{deva11} is a set of
non-oriented bounding boxes. We transform them to the oriented ones using the
online code that~\cite{deva11} provides.}. This step allows us to obtain a
manageably sized state space and simplifies the representation of a given
instance.
Suppose there are $m$ human parts in total ($m=6$ in this work), and then each image
is represented by $m$ candidate ensembles, each of which contains
$K_i$ candidates respectively.
Thus, the input space (i.e., sample space) $\mathcal{X}$ of our approach is
defined as
\begin{equation}
\mathcal{X} = \{ \x \mid \x =(\b_1, \b_2, \cdots, \b_m) \},
\label{eq:X}
\end{equation}
where $\mathbf{x}$ refers to an image and $\b_i$ denotes the candidate ensemble for the $i$-th human part (there are $K_i$ candidates in $\b_i$). Furthermore, we introduce the following notation:
\begin{equation}
\mathcal{P} = \{\p \mid \p =(p_1, p_2, \cdots, p_m), \forall i, 1 \leq p_i \leq K_i\},
\end{equation}
where $p_i$ is a positive integer that indicates the index of the
candidate for the $i$-th human part. In this way, the task of HPE is formulated
as the problem of learning a prediction function from $\mathcal{X}$ to
$\mathcal{P}$.

\begin{figure}
\centering
\includegraphics[width=0.95\linewidth]{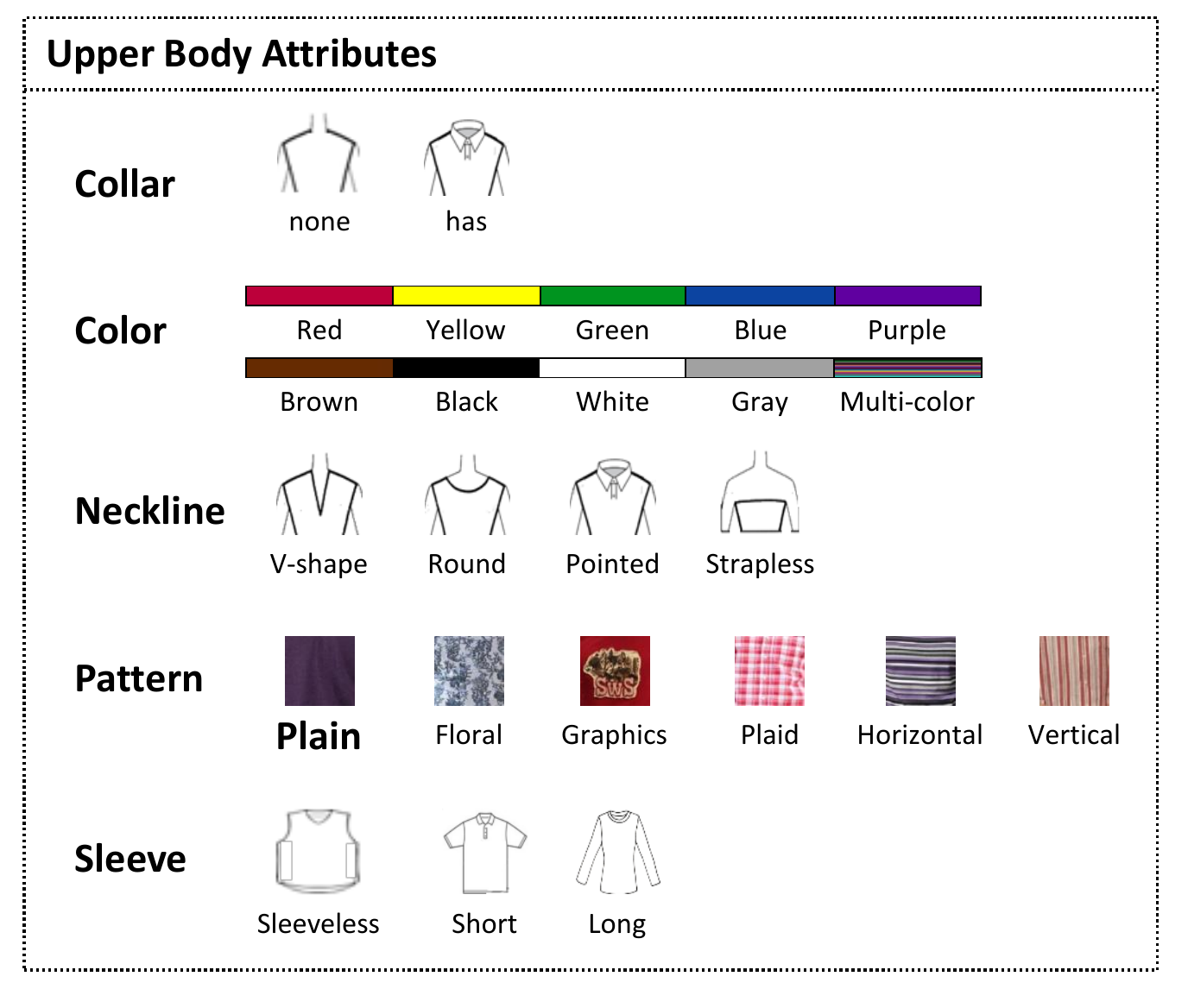}
\caption{
\textbf{Garment attributes definition.} We list all the garment attributes and types. In this work, we focus on the upper body dressing style. Part of the icons in this figure are quoted from~\cite{clothliu}.}
\label{fig:attr-def}
\end{figure}

\begin{table}
\caption{
\MakeLowercase{The interdependency between human parts and garment attributes.
First column: a garment attribute; Second column: the human parts associating with
the garment attribute shown in the first column; Third column: the
corresponding feature descriptors used to describe the parts (or attribute).}}

\centering
\begin{tabular}{|l|c|c|}
\hline
Attribute & Human Parts & Low-level Features\\
\hline
\hline
Collar    & Torso$+$Head & HOG\\
\hline
Color     & Torso       & Color Histogram\\
\hline
Neckline  & Torso$+$Head & HOG\\
\hline
Pattern   & Torso       & LBP~\cite{lbp}\\
\hline
Sleeve    & All arms    & Color Histogram\\
\hline
\end{tabular}
\label{tab:cloth-dep}
\end{table}

The goal of GAC in our work is to determine the garment attributes possessed
by each image. We consider five
types of attributes, including ``Collar'', ``Color'', ``Neckline'', ``Pattern'' and ``Sleeve''
 that are most relevant with the upper body limbs.
Each attribute has multiple styles, e.g., short sleeve and
long sleeve for the ``Sleeve'' attribute. We use $T_k$ to denote the number of attribute values for the $k$-th attribute.
The attribute values we consider in this paper are given in Figure~\ref{fig:attr-def}.
For the ease of presentation, we introduce the following notation:
\begin{equation}
\mathcal{C} = \{\c \mid \c =(c_1, c_2, \cdots, c_n), \forall k, 1 \leq c_k \leq T_k\},
\end{equation}
where $n$ is the number of garment attributes ($n=5$ in this work), and $c_k$
is the label for the $k$-th attribute (e.g., $c_5 = 1$ means short sleeve, and $c_5 = 2$
means long sleeve). In this way, similar with HPE, the task of GAC can also be formulated as
a problem of learning a prediction function from $\mathcal{X}$ to $\mathcal{C}$.

Hence, the task of performing combined HPE and GAC can be formulated as follows:
\begin{equation}
f: \mathcal{X} \rightarrow \mathcal{Y},
\label{eq:f}
\end{equation}
where $\mathcal{Y}$ is the joint output space defined by
\begin{equation}
\mathcal{Y} = \{\y \mid \y = (\p, \c), \p \in \mathcal{P}, \c \in \mathcal{C}\}.
\end{equation}

Regarding the prediction function $f$, we first presume that there is a compatibility function $S$
that measures the fitness between an input-output pair $(\x, \y)$:
\begin{equation}
S(\x, \y; \w) = \mathbf{w} \cdot J(\x, \y) + \alpha Q(\x, \p),
\label{eq:F(x,y;w)}
\end{equation}
where $\mathbf{w} \cdot J(\x, \y)$ denotes the inner product of two vectors,
$J(\x, \y)$ is the joint feature representation (which should be designed
carefully), $\w$ is an unknown weight vector (which should be learned
from training samples), $Q(\x, \p)$ is the energy function indicating the
response of a strong edge around the potential human parts and $\alpha$ is a parameter (which
can be hand-tuned by cross-validation).

In this way, the mapping function $f$ in Eq.~\eqref{eq:f} can be written as:
\begin{equation}
f(\x; \w) = \argmax_{\y \in \mathcal{Y}} S(\x, \y; \w).
\label{eq:f(x;w)}
\end{equation}

\subsection{Joint Feature Representation}\label{subsec:joint-feature}
The joint feature representation $J(\x, \y)$ is an important
component of the prediction function. In our approach, $J(\x, \y)$ consists of three types of features, including the
\emph{pose-specific} features denoted by $J_p(\x, \p)$, the
\emph{garment-specific} features denoted by $J_c(\c)$, and
the \emph{cross-task} features denoted by $J_{pc}(\x,\y)$; that
is,
\begin{equation}
\w \cdot J(\x,\y)  =  \w_p \cdot J_p(\x, \p)  +  \w_c \cdot J_c(\c) +  \w_{pc} \cdot J_{pc}(\x,\y).
\label{eq:joint-feature}
\end{equation}

In the following, we shall present our techniques used to design each type of
feature.
\subsubsection{Pose-Specific Features} \label{subsubsec:pose-feature}
\begin{figure}
\centering
\includegraphics[width=0.5\linewidth]{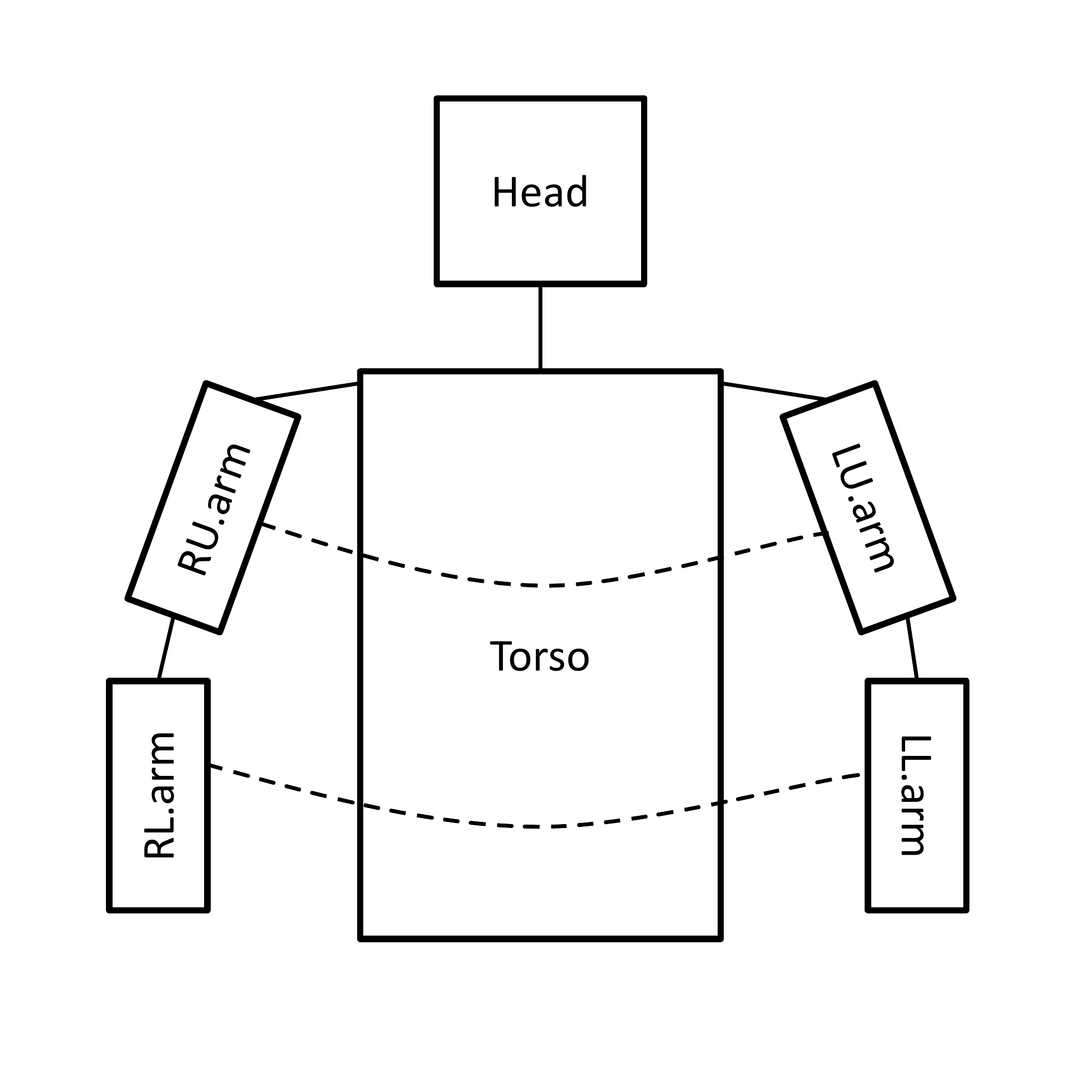}
\caption{\textbf{Part-part relations.} The solid lines are used to indicate
spatially connected parts, e.g., torso and right upper arm (RU.arm). We mainly
consider the geometry constraint for this case, e.g., relative position and
relative rotation. The dashed
lines mean the relations between symmetric parts, e.g., right upper arm
(RU.arm) and left upper arm (LU.arm). For these parts, we model the appearance
constraint, i.e., appearance similarity in color and texture descriptors.}
\label{fig:model:p}
\end{figure}
Given an image represented as $m$
candidate ensembles (each candidate is
a bounding box), we extract some features
specifically useful for HPE, called pose-specific features, as follows:
\begin{equation}
\begin{split}
\w_p \cdot J_p(\x, \p) =& \sum_{i=1}^{m} \w_p^{u,i} \cdot \phi_p(\x, p_i)  \\
& + \sum_{(i,j)\in E_p^d} \w_p^{d, ij} \cdot \psi_p^d(\x, p_i, p_j)  \\
& + \sum_{(i,j)\in E_p^c} \w_p^{c, ij} \cdot \psi_p^c(\x, p_i, p_j),
\end{split}
\end{equation}
where $\phi_p(\x, p_i)$ denotes the unary feature for the $i$-th human
part, $\psi_p^d(\x, p_i, p_j)$ models the pairwise relations between spatially connected human parts,
$E_p^d$ is the collection of all pairs of connected parts,
$\psi_p^c(\x, p_i, p_j)$ contains the appearance consistency message between symmetric parts
and $E_p^c$ is the set of all symmetric parts (see Figure~\ref{fig:model:p} for details). The three terms are called \emph{unary score}, \emph{deformation score} and \emph{consistency score} respectively.

In our approach, the unary feature
$\phi_p(\x, p_i)$ is chosen as the HOG descriptor~\cite{hog} which has been proved quite effective for object detection~\cite{dpm, deva11}.

The design of $\psi_p^d(\x, p_i, p_j)$ concerns some basic geometry constraints between connected parts, including relative position, rotation and distance of part candidate $p_i$ with respect to $p_j$. More concretely, we divide the image space into 3 by 3 regions, with $p_j$ at the central region. Then we use a $9$ dimensional one-zero vector as the relative position feature, where there is only one element with value ``1'' that indicates the region where $p_i$ is located. To describe the relative rotation, we divide the range of angles $[0, 360]$ into 20 bins and use a $20$ dimensional one-zero vector as the feature. Our relative distance feature is the euclidean distance between the center of $p_i$ and $p_j$.

For the symmetric parts, we assume some consistency constraints between them which hold for most cases; that is, they should share similar appearance. In this work, we compute the divergence of the color histogram in RGB and LAB space and take it as the feature descriptors.

In Figure~\ref{fig:model:p}, we mark all the parts that are related to each other.

\subsubsection{Garment-Specific Features}
There are some features only specific for garment, i.e., garment-specific features. In this work, we
consider the co-occurrences between different garment attribute values:
\begin{equation}
 \w_c \cdot J_c(\c) = \sum_{k,l} \w_c^{kl} \cdot \psi_c(c_k, c_l),
\label{eq:cloth-feature}
\end{equation}
where $\psi_c(c_k, c_l)$ is a binary vector that indicates whether or not $c_k$
and $c_l$ co-occur in an image. For example, the texture type ``drawing'' (usually
belongs to T-shirt style) often co-occurs with the collar type ``round''.

\subsubsection{Cross-Task Features}
The cross-task features encode the correlations between human parts and garment attributes.
In our approach, we model the part-garment relations manually specified as in Table~\ref{tab:cloth-dep}. For a given attribute $k$, we denote the human part(s) associated with it
as $\hat{\p}(k)$ and the corresponding configuration(s) as $\hat{\p}_{k}$. Then the cross-task features are formulated as:
\begin{equation}
 \w_{pc} \cdot J_{pc}(\x,\y) = \sum_{k=1}^{n} \w_{pc}^{k} \cdot \Psi_{pc}^{k} (\x, \hat{\p}_{k}, c_k),
\label{eq:cross-feature}
\end{equation}
where $\Psi_{pc}^{k}(\x, \hat{\p}_k, c_k)$ denotes the features extracted from $\x$ under the constraints of part configuration $\hat{\p}_k$ and attribute label $c_k$. Note that here we write the cross-task score as a summary by the attribute order. Since the dependency between part and attribute is cyclic, one can also write it by the human part order.

To describe the design of $\Psi_{pc}^k (\x, \hat{\p}_k, c_k)$, we first convert the garment attribute label $c_k$ to a $T_k$ dimension vector $\mathrm{I}(c_k)$, with only one dimension assigned with value one and others with zeros.
From Table~\ref{tab:cloth-dep}, the low-level feature descriptors of the $k$-th
garment attribute depend on two aspects: the corresponding human part(s)
and the feature type (denoted by $F_k$ and specified in Table \ref{tab:cloth-dep}).
We use $F_k(\hat{\p}_k)$ to denote features of the $k$-th garment attribute under
the part candidate(s) $\hat{\p}_k$.
Then our cross-task feature $\Psi_{pc}^k(\x,\hat{\p}_k, c_k)$ is represented as
follows:
\begin{equation}
\Psi_{pc}^k(\x, \hat{\p}_k, c_k) = F_k(\hat{\p}_k) \otimes \mathrm{I}(c_k),
\end{equation}
where the ``$\otimes$'' operator is the outer product of two vectors. In fact,
we map the resulting matrix to a vector by the row order.
Note that in Table~\ref{tab:cloth-dep},
a garment attribute depends exclusively on the some of the limbs, not all ones.
This technique that feature descriptors draw from both
the labels of human parts and garment attributes, provides us a simple way to
capture the correlations between HPE and GAC and makes it a unified approach
towards the two intertwined problems.

\subsection{Learning with Structured SVM}
We perform our joint estimation for HPE and GAC using the prediction function
$f$ in Eq.~\eqref{eq:f(x;w)}.
The weight vector $\w$ is a critical component of the prediction
function. Given $N$ training samples $\{(\x_r,
\y_r)\}_{r=1}^N$, we compute $\w$ by solving the following
structured SVM problem:
\begin{equation}
\begin{split}
\minimize_{\w, \xi}\ &\frac{1}{2}\Vert \w \Vert^2 + C\sum_{r=1}^N\xi_r,\\
\textrm{subject to}\ &\forall r \in \textrm{pos},  \w \cdot J(\x_r, \y_r) \geq  1-\xi_r, \\
&\forall r \in \textrm{neg},  \w \cdot J(\x_r, \y_r) \leq  -1+\xi_r.\\
\end{split}
\label{eq:svm-form}
\end{equation}
where $C$ is the parameter that controls the trade-off between margin and accuracy, and
$\xi_r\geq0$ is a slack variable. In our experiments, we set $C$ with a fixed value $0.01$ that
allows a soft margin.

\begin{figure*}
\centering
\subfloat[]
{
\includegraphics[width=0.16\linewidth]{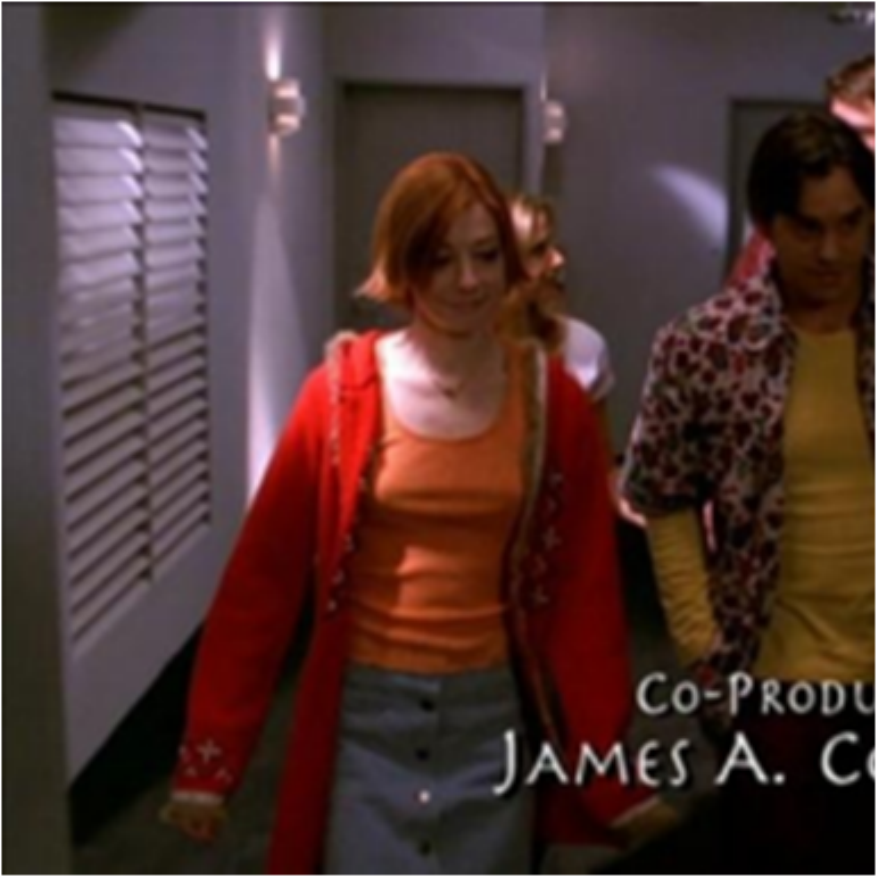}
}
\subfloat[]
{
\includegraphics[width=0.16\linewidth]{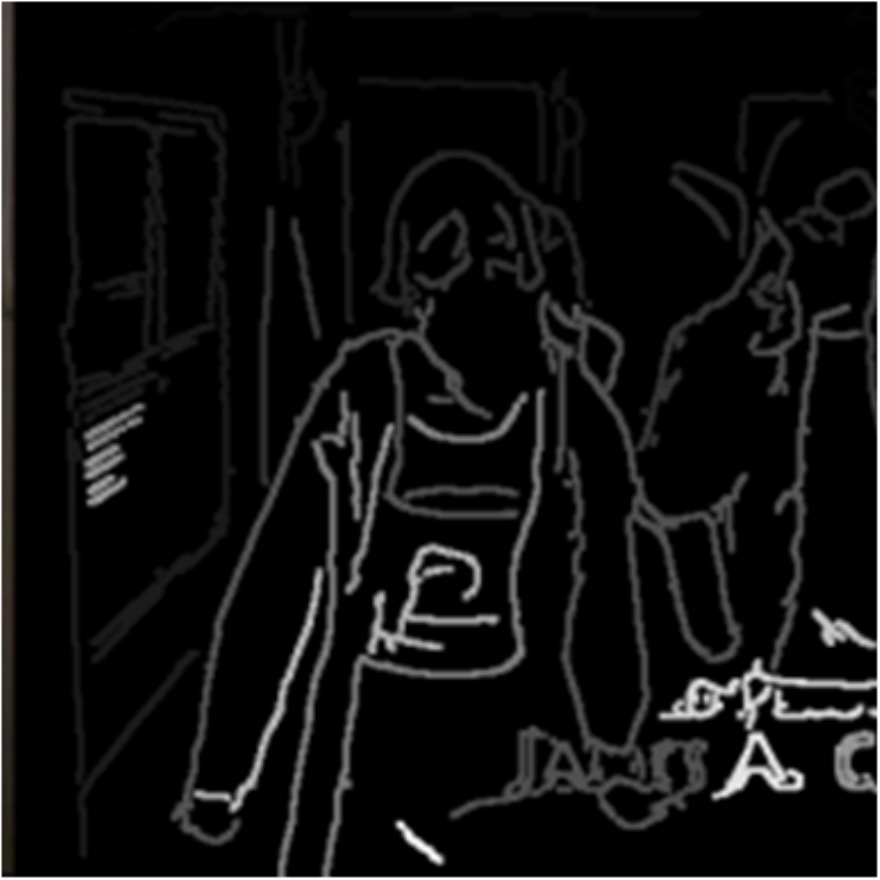}
}
\subfloat[]
{
\includegraphics[width=0.16\linewidth]{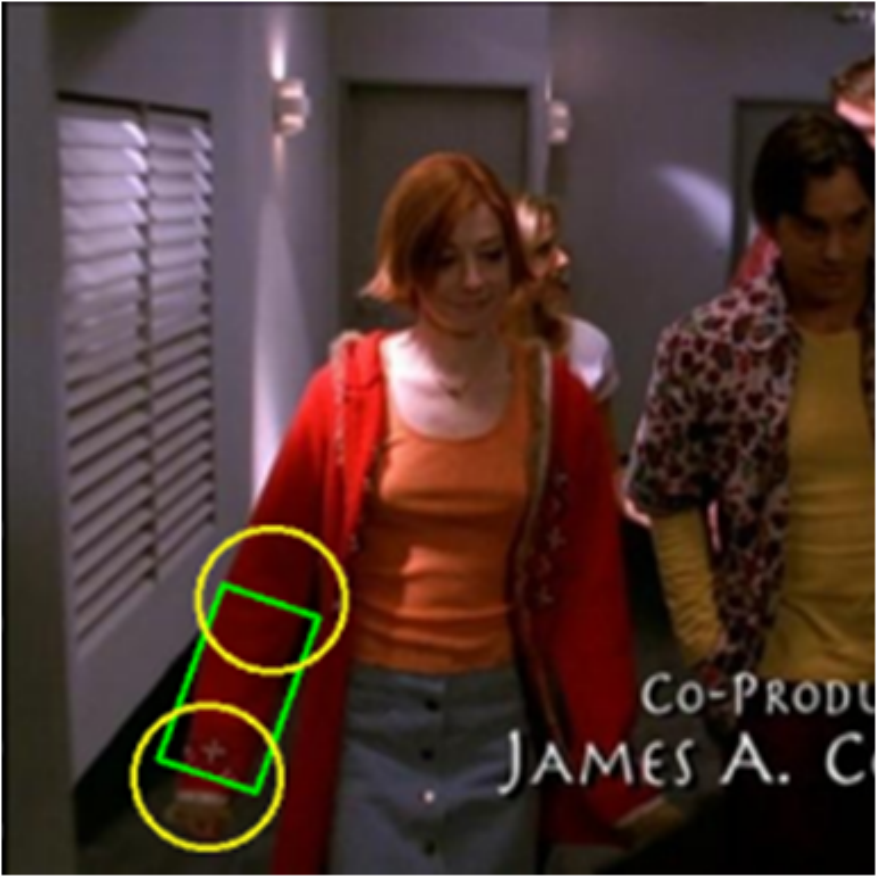}
}
\subfloat[]
{
\includegraphics[width=0.16\linewidth]{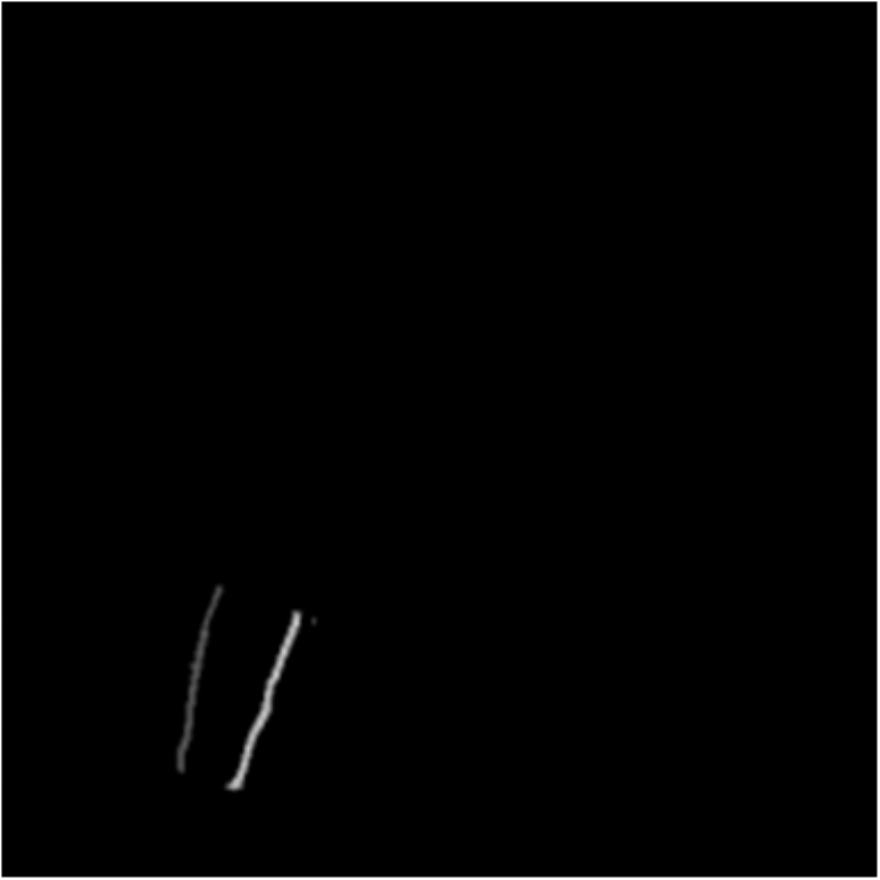}
}
\subfloat[]
{
\includegraphics[width=0.16\linewidth]{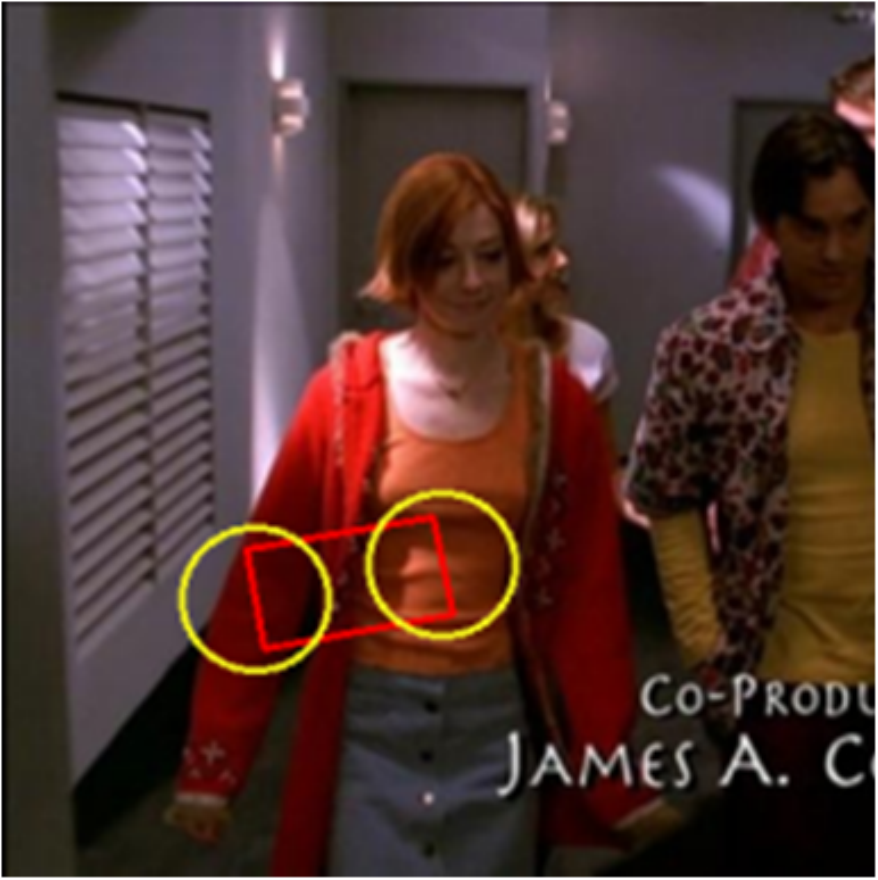}
}
\subfloat[]
{
\includegraphics[width=0.16\linewidth]{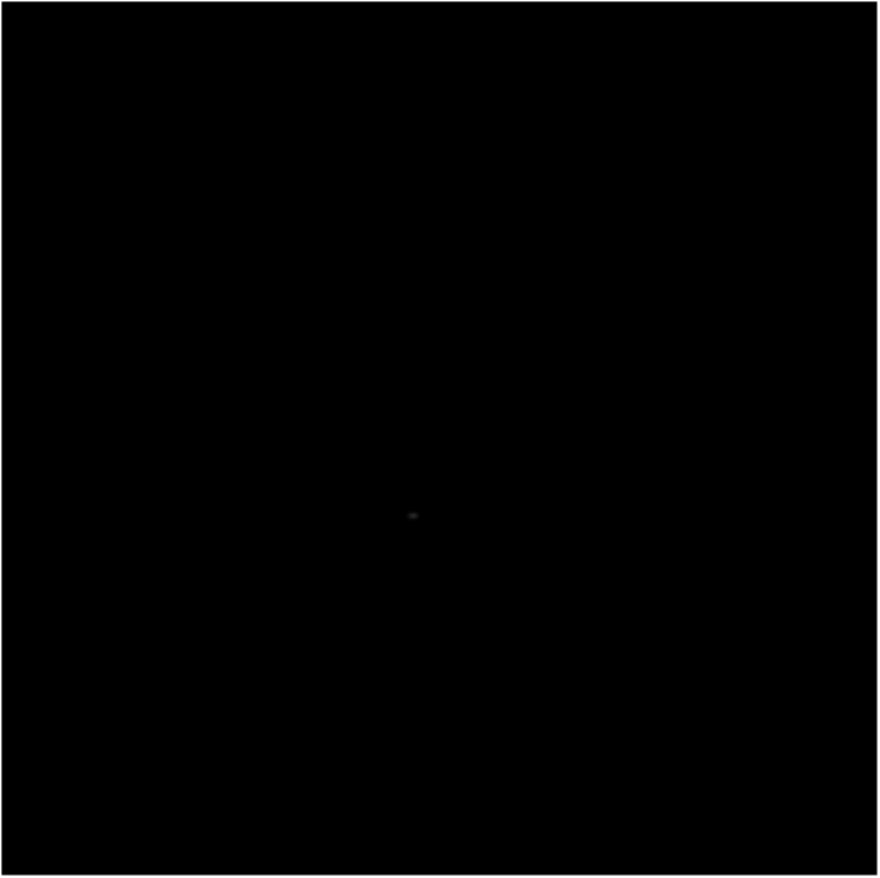}
}
\caption{\textbf{Illustration for Strong Edge Evidence.} (a) original image; (b) boundary detection results; (c) a correct candidate for the right-lower arm (green rectangle); (d) strong edge evidence for the correct candidate; (e) an incorrect prediction for the same arm (red rectangle); (f) strong edge evidence for the incorrect candidate. Given a part candidate (denoted by an oriented rectangle), we extract the strong edges that connect the regions of the joints of the part (denoted by two yellow rounds).}
\label{fig:sedge}
\end{figure*}

\subsection{Strong Edge Evidence} \label{subsec:sedge}
Now we give details on the design of our energy function $Q(\x, \p)$ in Eq.~\eqref{eq:F(x,y;w)}. First, we utilize a boundary detector~\cite{gpb} to detect all potential edges in an image. Then for a candidate human part, we try to find the long edges that connect the two joint regions as the strong edge evidence (see Figure~\ref{fig:sedge}).
Our energy function mainly considers three factors: the consistency of orientation between the part $p_i$ and the strong edges $\mathbf{se}_i$ (denoted by $Q^o(p_i, \mathbf{se}_i)$),
the distance of the part away from the strong edges (denoted by $Q^d(p_i, \mathbf{se}_i)$) and the strength of the strong edges themselves. That is,
\begin{equation}
Q(\x, \p) = \sum_{i=1}^m Q^o(p_i, \mathbf{se}_i) + \beta \sum_{i=1}^m Q^d(p_i, \mathbf{se}_i),
\label{eq:E(x,p)}
\end{equation}
where $\beta$ is a parameter which can be tuned by cross validation. Given a part candidate $p_i=(x,y,\theta,s)$, the first term in Eq.~\eqref{eq:E(x,p)} is computed as:
\begin{equation}
Q^o(p_i, \mathbf{se}_i) = \frac{1}{Z}\sum_{e \in \mathbf{se}_i}\cos(\theta-\theta_e)\cdot strg_e,
\end{equation}
where $e$ is an image pixel on the strong edges $\mathbf{se}_i$, $Z$ is the number of pixels on $\mathbf{se}_i$, $\theta_e$ is the orientation of the strong edge at pixel $e$ and $strg_e$ is the edge strength (which is produced by the algorithm~\cite{gpb}). The measurement of the distance from the given part to the strong edges is represented as follows:
\begin{equation}
Q^d(p_i, \mathbf{se}_i) = \frac{1}{Z}\sum_{e \in \mathbf{se}_i} \min\{ \mathrm{dt}(e, l), \mathrm{dt}(e,r) \}\cdot strg_e,
\end{equation}
where $l$ and $r$ are the two parallel edges of the part bounding box whose angles are $\theta$, and
$\mathrm{dt}(e,l)$ is the closest distance of pixel $e$ from edge $l$ which can be efficiently computed by a distance transform algorithm~\cite{dt}. Note that if there is no strong edge associated with the underlying part, we force the energy to be zero.

\subsection{Inference} \label{subsec:inference}
\begin{algorithm}
\caption{Approximate Inference for Joint Estimation}
\label{alg:whole}
\begin{algorithmic}[1]
    \REQUIRE An input sample $\x$, the weight vector $\w$, parameter $\alpha$ and $\beta$.
    \ENSURE Optimal joint estimation $\y^*$.
    \STATE Set the optimal joint estimation $\y^*=\varnothing$.
    \STATE Set the optimal score $S^*=-\infty$.
    \STATE Initialize the parts estimation: \\
    $\p_0 = \argmax_{\p\in\mathcal{P}} \w_p \cdot J_p(\x, \p) + \alpha Q(\x, \p)$.
    \REPEAT
        \STATE Compute the garment attributes:\\
               $\c_t = \argmax_{\c \in \mathcal{C}}  \w_c \cdot J_c(\c) +  \w_{pc} \cdot J_{pc}(\x, \p_{t-1}, \c)$.
        \STATE Compute the parts estimation:\\
               $\p_t = \argmax_{\p \in \mathcal{P}} \w_p \cdot J_p(\x, \p) + \alpha Q(\x, \p)
               +  \w_{pc} \cdot J_{pc}(\x, \p, \c_t)$.
        \STATE Compute the local score:
               $S = S(\x, \y_t; \w)$.
        \IF{$S>S^*$}
           \STATE $S^*=S$, $\y^*=\y_t$.
        \ENDIF
    \UNTIL{$S^*$ not change}
\end{algorithmic}
\end{algorithm}
Now that we have clarified how to design the joint feature, the strong edge energy function, as well as the learning algorithm for weight vector $\w$ in Eq.~\eqref{eq:F(x,y;w)}, we propose our inference algorithm which is quite efficient (for each input sample, our algorithm only needs 2 seconds for the joint estimation) and effective.

\begin{figure}
\centering
\includegraphics[width=\linewidth]{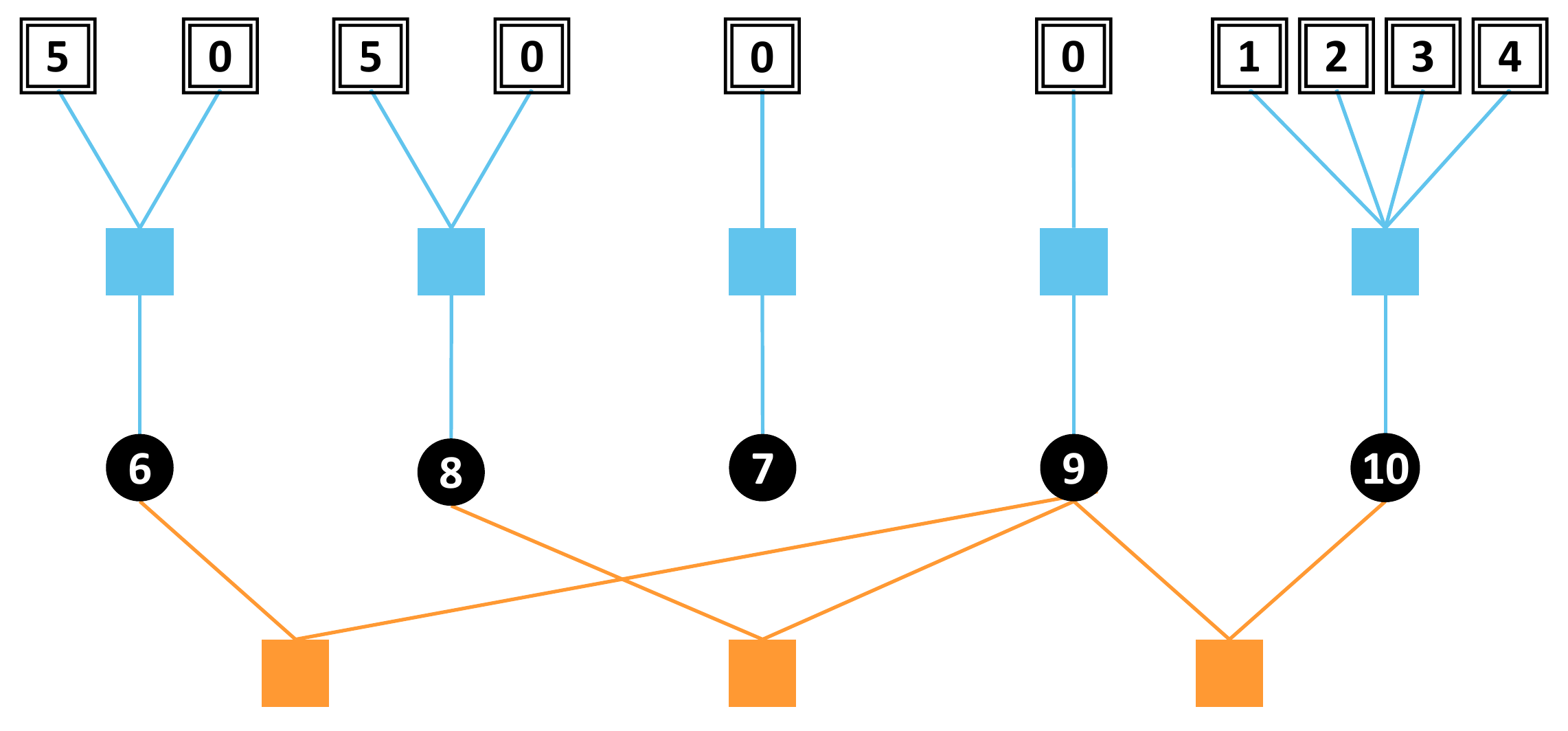}
\caption{\textbf{The factor graph representation for inferring attributes.} Part nodes are with fixed values and stretching them will not affect the optimal solution. Unary, deformation and consistency potentials of part nodes are not drawn here, as they don't contribute to searching for a best attribute solution.}
\label{fig:infer-for-attr}
\end{figure}

In Figure~\ref{fig:whole-model}, we represent our problem as a factor graph $\mathcal{G}$, where the black-rectangle node denotes a human part, the black-circle node denotes a garment attribute and the colored node denotes a potential. As our original problem is a cyclic graph, it cannot be optimized exactly and efficiently. Therefore, in Algorithm~\ref{alg:whole}, we propose an iterative algorithm to search for an approximate solution. Our algorithm receives a sample $\x$ (defined in Eq.~ \eqref{eq:X}), the SVM weight $\w$, parameter $\alpha$ and $\beta$ as inputs and outputs the optima for the joint problem. In each iteration, by fixing one type of the variable (either human part or garment attribute, see step $5$ and step $6$), our inference procedure can be performed on a tree structure which yields an efficient computation by dynamic programming~\cite{ps2}. This procedure is also illustrated in Figure~\ref{fig:infer-for-pose} and Figure~\ref{fig:infer-for-attr}.

\subsubsection{Inference for Pose} \label{subsubsec:infer-for-pose}
\begin{figure}
\centering
\includegraphics[width=\linewidth]{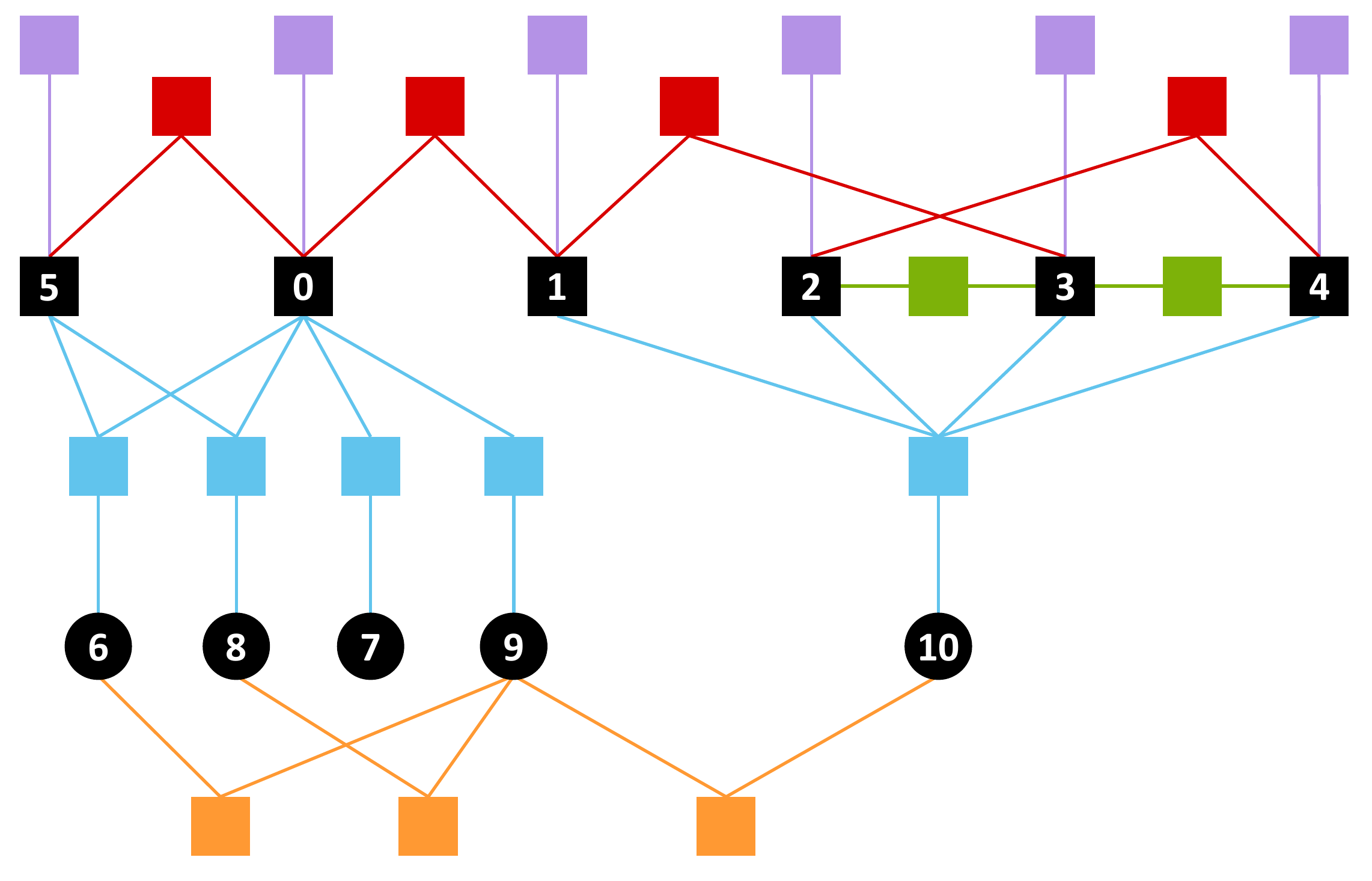}
\caption{\textbf{The factor graph representation for our problem.} We denote our variables with black nodes, those of which with number $0$--$5$ represent the human parts: torso, RU/LU/RL/LL arm and head, while those with number $6$--$10$ denote the garment attributes: collar, color, neckline, pattern and sleeve. We denote our potentials with colored nodes, with purple ones denoting the unary potential, red denoting the deformation potential, green denoting the consistency potential, orange denoting the attribute co-occurrence potential and cyan denoting the cross potential.}
\label{fig:whole-model}
\end{figure}

\begin{figure}
\centering
\includegraphics[width=\linewidth]{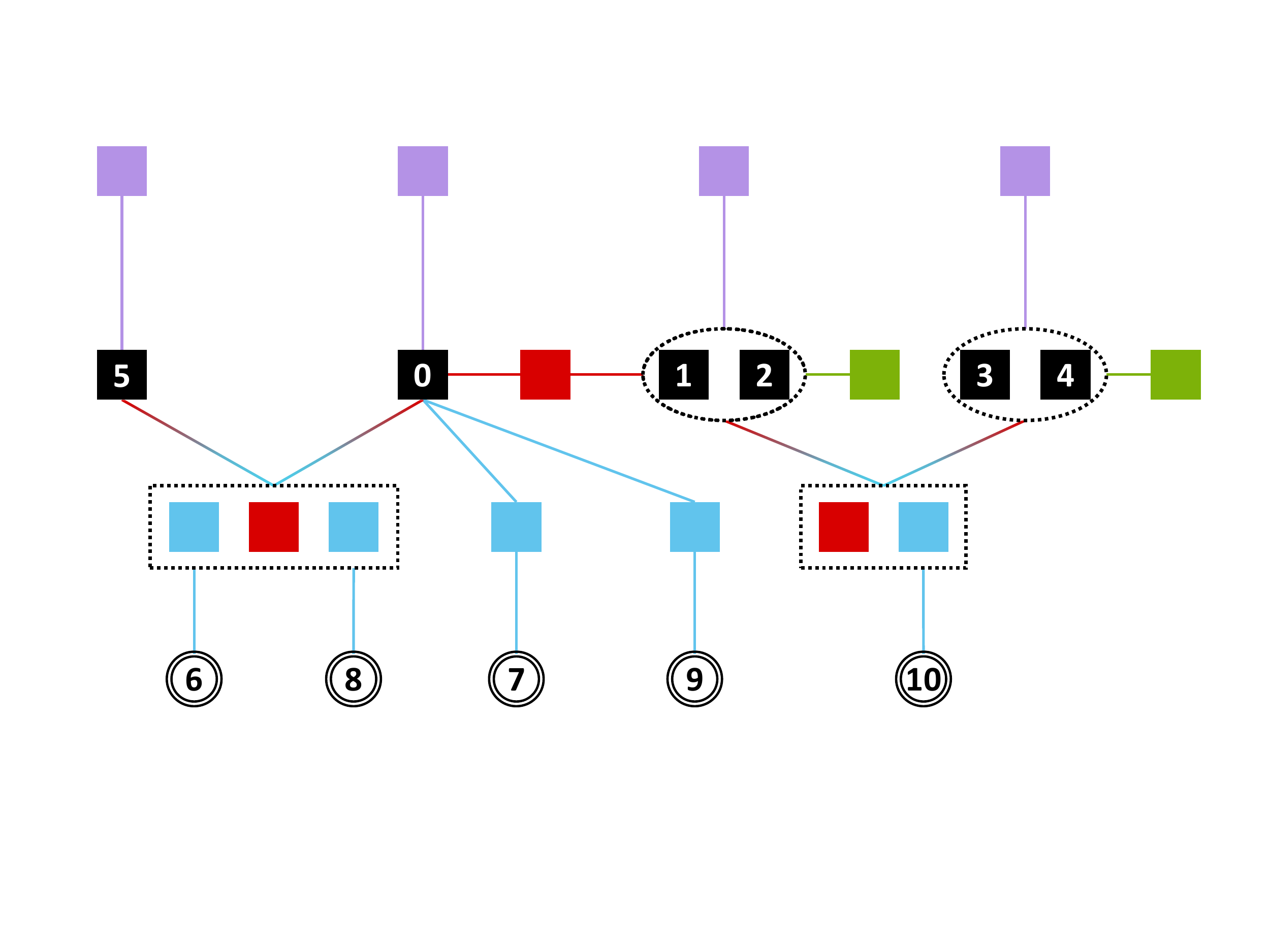}
\caption{\textbf{The factor graph representation for inferring human pose.} Circle nodes with double boundaries are assigned with fixed values. Symmetric parts are grouped into a super-node, denoted by a dashed oval. For some part nodes, their deformation and cross potential can also be grouped as the associated attribute nodes are now fixed. Note that we don't draw the garment-specific potentials as they don't contribute for searching a best pose.}
\label{fig:infer-for-pose}
\end{figure}

In the work of~\cite{ps2}, the PS model is restricted as a tree: each node has a unary term that describes how suitable a configuration is assigned to this part, and each edge encodes the deformation cost for a pair of connected parts. In Figure~\ref{fig:whole-model}, we demonstrate our extension for the traditional PS model:
\begin{itemize}
\item appearance consistency between symmetric parts (green nodes)
\item joint compatibility across the human part(s) and the garment attribute(s) (blue nodes)
\end{itemize}
Adding the edges connecting symmetric parts will destroy the tree structure. In Figure~\ref{fig:infer-for-pose}, however, we propose a trick to group the symmetric parts as a \emph{super-node} so that the global structure remains to be a ``tree''. On the other hand, an edge across human part and garment attribute is used to measure how compatible a human part configuration is with a given attribute. We call this kind of cost as \emph{cross score}. When the attribute variables are fixed, we can remove the garment-specific potentials as they do not contribute to searching for the best pose. In addition, we can group some deformation and cross potentials for a more concise representation, e.g. as what we do for node 1 and 2 in Figure~\ref{fig:infer-for-pose}.

In Algorithm~\ref{alg:infer-for-pose}, we describe our computation procedure.
For a super-node $\mathbf{i}$~\footnote{For simplicity, here we call all variable nodes as super-nodes.}, we denote its children nodes as $C_{\mathbf{i}}$.
In the line $3$--$13$, we first compute the scores with respect to a single node $\mathbf{i}$. This step involves calculation of unary score, strong edge score, consistency score and cross score. Note that we have grouped the symmetric parts as one node. In this way, the consistency score is a self description towards the node $\mathbf{i}$. In line $15$, we compute the deformation score of node $\mathbf{i}$ and $\j$. For example, if the super-node $\mathbf{i}=\{1, 2\}$ whereas the super-node $\j=\{0\}$, the deformation score between $\mathbf{i}$ and $\j$ is the sum of deformation score of $(1, 0)$ and $(2, 0)$. In line $16$, we compute the cross score for all attributes whose associated human parts are exactly $\mathbf{i} \cup \j$. For example, the attribute node $6$ is associated with part node $\mathbf{5}$ and $\mathbf{0}$, so we will compute the cross score with respect to nodes $\{\mathbf{5}, \mathbf{0}, 6\}$. Line $18$--$27$ is a conventional message passing procedure that can be computed efficiently by dynamic programming~\cite{ps2}.
\begin{algorithm}
\caption{Exact Inference for Human Pose Estimation (Extended Pictorial Structure Inference)}
\label{alg:infer-for-pose}
\begin{algorithmic}[1]
    \REQUIRE An input sample $\x$, the weight vector $\w$, parameter $\alpha$, $\beta$ and garment attributes $\c$.
    \ENSURE Optimal pose estimation $\p^*$.
    \STATE Set the optimal joint estimation $\p^*=\varnothing$.
    \STATE Set the node $0$ as the root node.
    \FOR{each configuration $\p_{\i}$ of super-node $\i$}
    \STATE
           $\mathrm{m}_1 = \sum_{i \in \i} \w_p^{u,i} \cdot \phi_p(\x, p_i)$.
    \STATE
           $\mathrm{m}_2 = \alpha\sum_{i \in \i} Q(\x, p_i)$.

    \IF{$\i \in E_p^c$}
    \STATE
           $\mathrm{m}_3 = \w_{p}^{s,\i} \cdot \psi_p^s(\x, \p_{\i})$,
    \ELSE
    \STATE  $\mathrm{m}_3 = 0$.
    \ENDIF

    \STATE $\mathrm{m}_4 = \sum_{k, \hat{\p}(k) = \i} \w_{pc}^k \cdot \Psi_{pc}^k(\x, \hat{\p}_k, c_k)$.

    \STATE set $\mathrm{m}(\p_{\mathbf{i}}) = \mathrm{m}_1 + \mathrm{m}_2 + \mathrm{m}_3 + \mathrm{m}_4$.
    \ENDFOR
           \FOR{each configuration of parent-child pair $\p_{\j}$ and $\p_{\i}$}
           \STATE 
                  $\mathrm{l}_1 = \sum_{i \in \mathbf{i}, j \in \j, (i, j)\in E_p^d} \w_p^{d,ij} \cdot \psi_p^{d}(\x, p_i, p_j)$.
           \STATE $\mathrm{l}_2 = \sum_{k, \hat{\p}(k)= \i \cup \j} \w_{pc}^k \cdot \Psi_{pc}^k(\x, \hat{\p}_k, c_k)$.
           \STATE set $\mathrm{l}(\p_{\i}, \p_{\j}) = \mathrm{l}_1 + \mathrm{l}_2$.

           \IF{$\i$ is a leaf node}
           \STATE $\mathrm{B}_{\i}(\p_{\j}) = \max_{\p_{\i}}\(\mathrm{m}(\p_{\i}) + \mathrm{l}(\p_{\i}, \p_{\j})\)$,
           \ELSE
           \STATE $\mathrm{B}_{\i}(\p_{\j}) = \max_{\p_{\i}}\(\mathrm{m}(\p_{\i}) + \mathrm{l}(\p_{\i}, \p_{\j}) + \sum_{\mathbf{v}\in C_{\i}}\mathrm{B}_{\mathbf{v}}(\p_{\i})\)$.
           \ENDIF
           \ENDFOR
    \STATE Compute the best configuration for the root node:\\
           $\p_{\mathbf{0}}^* = \argmax_{\p_{\mathbf{0}}}\(\mathrm{m}(\p_{\mathbf{0}}) + \sum_{\mathbf{v}\in C_{\mathbf{0}}}\mathrm{B}_{\mathbf{v}}(\p_{\mathbf{0}})\)$.
    \FOR{each parent-child pair $(\p_{\j}^*,\p_{\i})$}
    \STATE $\p_{\i}^* = \argmax_{\p_{\i}}\mathrm{B}_{\i}(\p_{\j}^*)$.
    \ENDFOR
\end{algorithmic}
\end{algorithm}

\subsubsection{Inference for Attributes} \label{subsubsec:infer-for-attr}
Referring to Figure~\ref{fig:infer-for-attr}, we stretch the part variables and remove redundant edges associated with the stretched variables from the original graph as they contribute nothing to this inference step. Note that for the attribute-attribute pairs (i.e. the garment-specific feature), we manually model them as a tree structure. In this way, we can still perform an efficient computation like Algorithm~\ref{alg:infer-for-pose}.

\subsubsection{Implementation and Computation Complexity} \label{subsubsec:comp-comp}
We write $K = \max_{1\leq i \leq m}K_i$ and $T = \max_{1\leq k \leq n} T_k$.
Now we propose a computation analysis for our Algorithm~\ref{alg:infer-for-pose} and give some optimization tricks. In line $3$--$13$, one has to loop over all possible configuration $\p_{\i}$ for the super-node $\i$, and there are at most $2$ nodes in a super-node (see Figure~\ref{fig:infer-for-pose}); this makes the computation $O(K^2)$.
In fact, note that the computation of unary and strong edge score can be \emph{decomposed} into a summation of each node (line $4$ and $5$), which indicates that we can separately compute these scores for each node configuration, yielding a computation $O(K)$. Also note that actually we only compute cross score for node $0$ in line $11$ (see Figure~\ref{fig:whole-model}). Based on this observation, computation on $m(\p_{\i})$ is reduced from $O(K^2)+O(K^2)+O(K^2)+O(K^2)$ to $O(K)+O(K)+O(K^2)+O(K)$.
 In line $14$--$16$, we compute the deformation and cross score for each pair $(\p_{\i}, \p_{\j})$, which yields a computation complexity $O(K^4)$ if without any optimization. For the deformation score, as the decomposition property still holds, the computation is $O(K^2)$. For the cross score in line $16$, when $k\in \{6, 8\}$, the computation is $O(K^2)$ since we have to loop over all the configurations for nodes ${0}$ and ${5}$. When $k=10$, however, it is not necessary to enumerate the $O(K^4)$ combinations of node $1$, $2$, $3$ and $4$ if we design a suitable cross-task feature. In our case, $F_{10}(\hat{\p}_{10})$ is the concatenation of the color histogram of each $p_i \in
\hat{\p}_{10}$, which implicitly owns the decomposition property. Thus, the computation can be reduced to $O(K)$, if we omit the summation operation of these separate scores.

\subsection{Parameter Sharing} \label{subsec:param-share}
Our work is distinct from other works which  address pose estimation and garment attribute in two aspects. First, our carefully designed structured learning model integrates the pose feature and garment attributes into a principle fashion, facilitating a global optimal model. Second, although we derive an iterative inference algorithm to approximate the optima, we allow the \emph{parameter sharing} between the two steps, i.e. the cross-task features are shared and contribute to both pose estimation and attribute classification (line $5$ and $6$ in Algorithm~\ref{alg:whole}). Therefore, our approach is a paradigm of learning globally and inferring locally, achieving both effectiveness and efficiency (see Section~\ref{sec:exp} for experimental justification).

\begin{figure}
\centering
\includegraphics[width=0.95\linewidth]{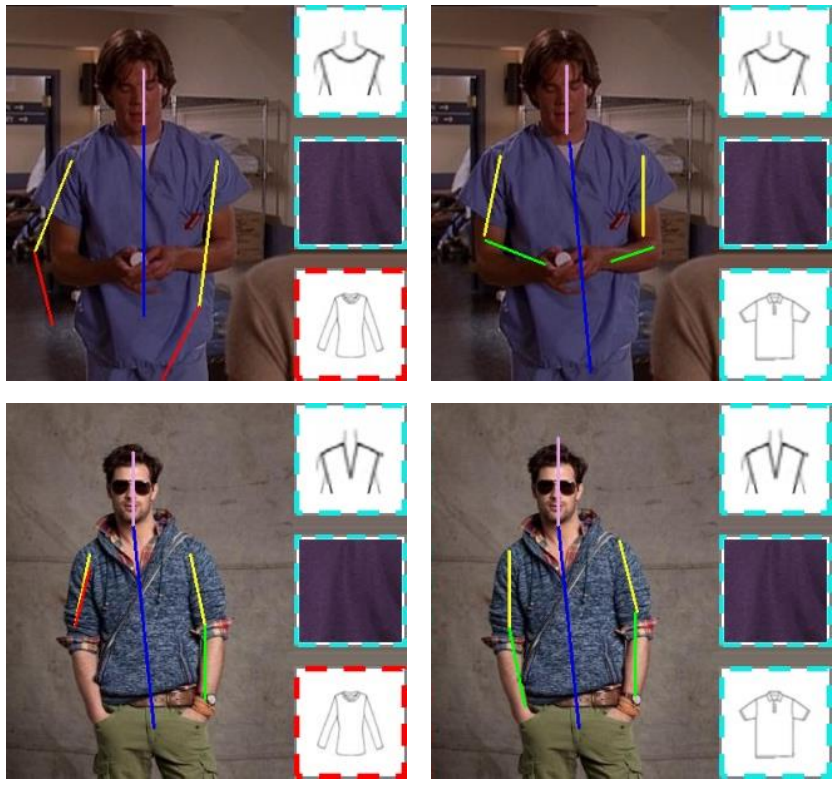}
\caption{
\textbf{Our joint approach v.s. YDR~\cite{deva11}}.
Left column: results from~\cite{deva11}. Right column: results
from our joint approach. \cite{deva11} estimates
incorrect lower arm(s), which subsequently results in an incorrect prediction for the
sleeve attribute. Our joint approach captures the co-relations between the
arms and the sleeve attribute and makes a correct estimation for \emph{both}.
}
\label{fig:deva-compare-example}
\end{figure}

\section{Experiments}\label{sec:exp}
\subsection{Experimental Settings}
In this section, we introduce our experimental settings, including the used
datasets, the baselines, the evaluation metrics and the scheme for training
structured SVM and inferring for a testing image.

\begin{figure*}
\centering
\subfloat[Buffy]
{
  \includegraphics[width=0.48\linewidth]{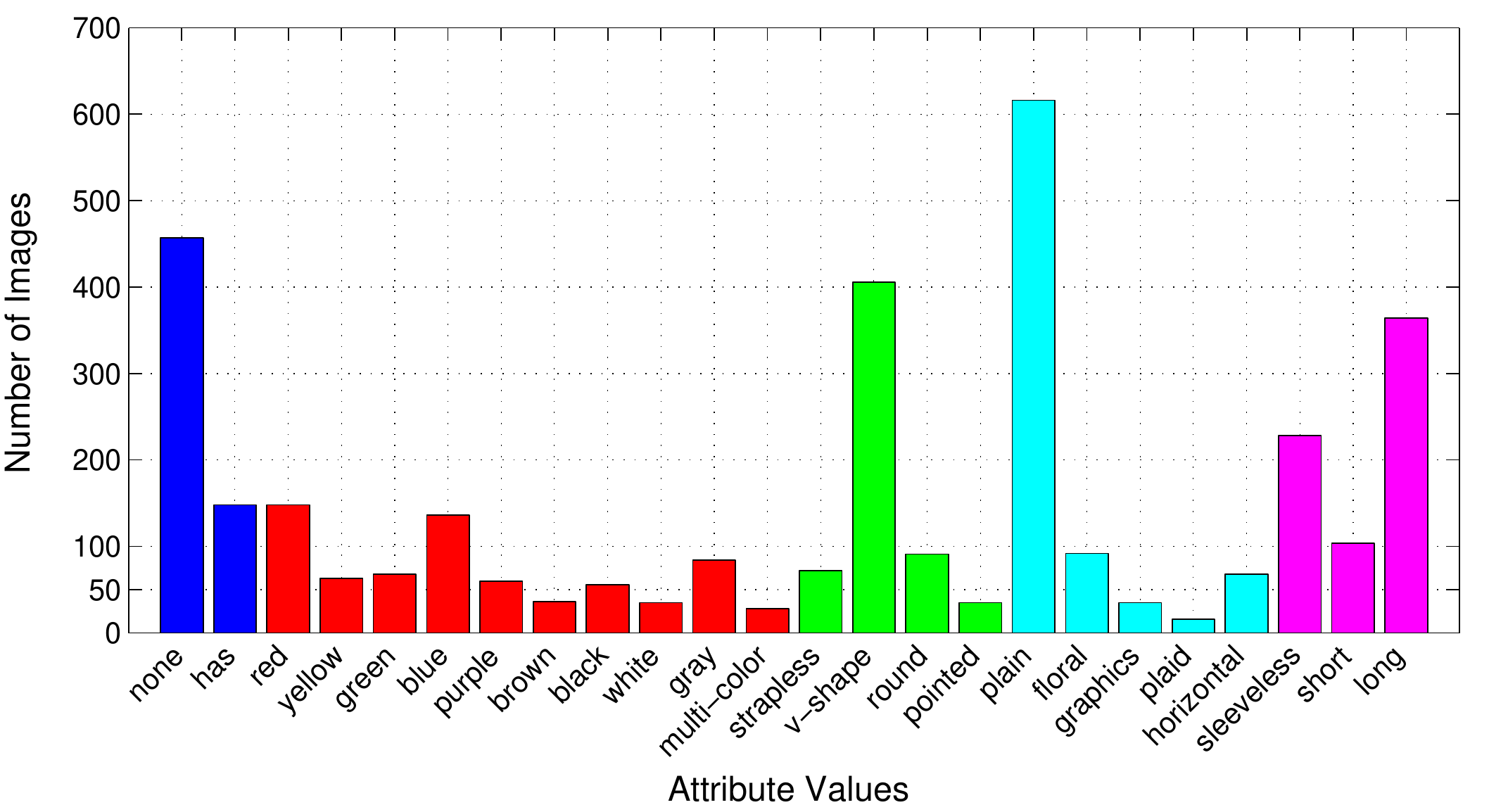}
}
\subfloat[DL]
{
  \includegraphics[width=0.48\linewidth]{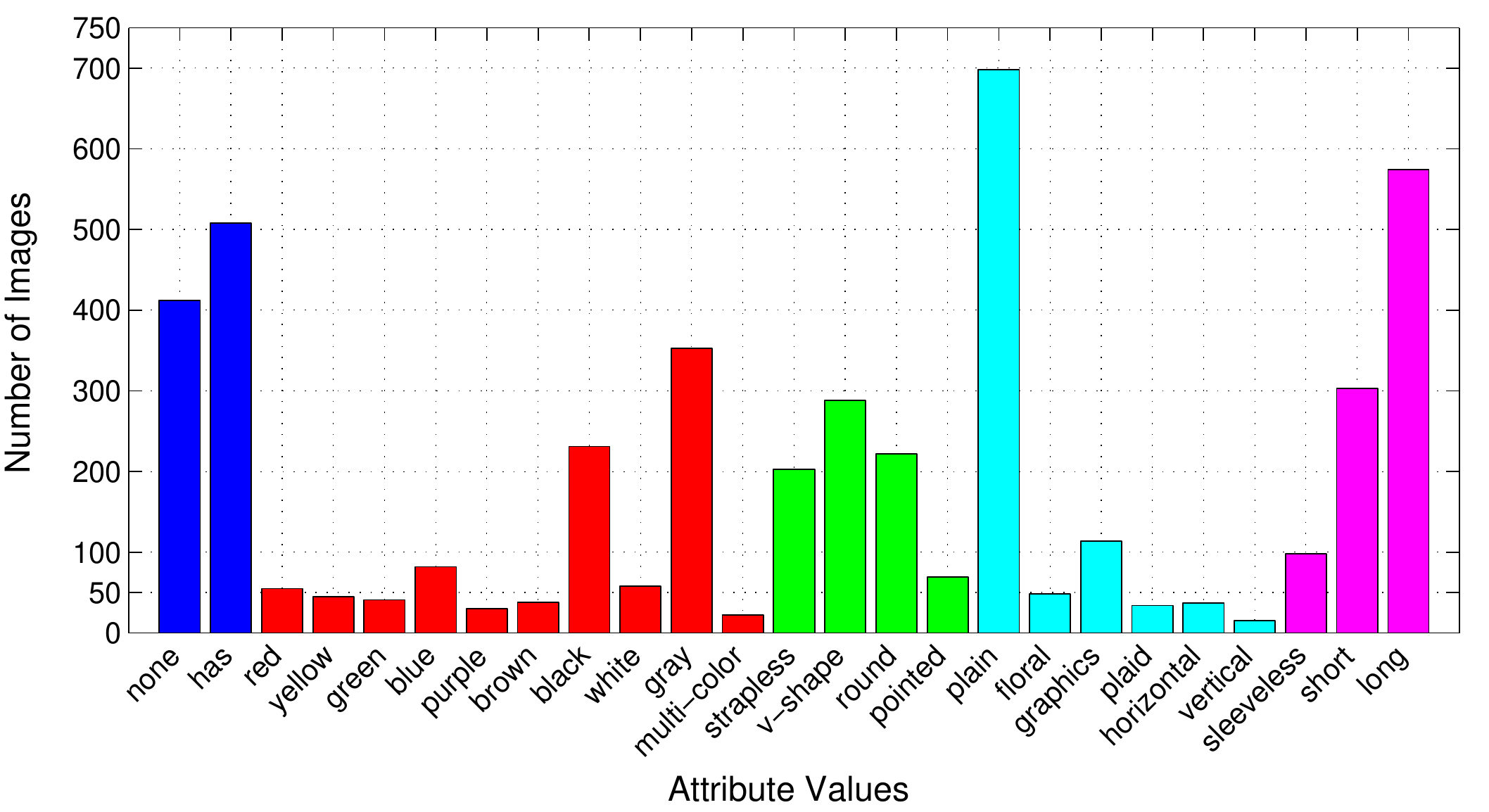}
}
\caption{Statistical information for the clothing attribute annotation. Left: Buffy dataset. Right: DL dataset. We use bars with different colors to denote different attributes. Orders from left to right defer to the orders in Figure~\ref{fig:attr-def}.}
\label{fig:cloth-stat}
\end{figure*}

\begin{figure}
\centering
\includegraphics[width=0.95\linewidth]{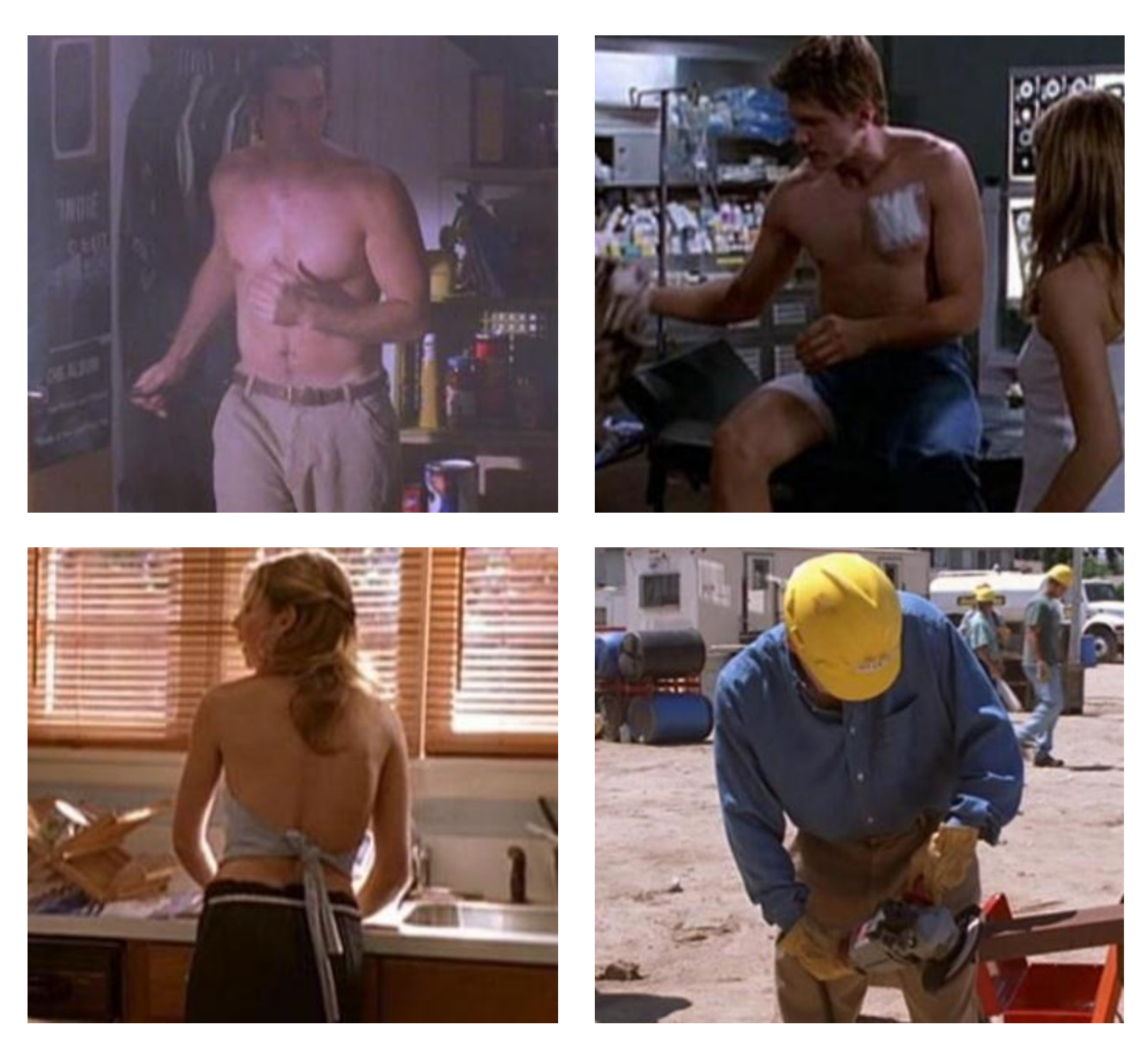}
\caption{Examples lacking some garment attributes. Characters
in these images all lack visual cues for specific garment attributes.
Persons of the first row wear no garment, thus cannot be labeled with any
garment attribute while those of the second row cannot be labeled with
part of attributes, e.g., neckline types.}
\label{fig:hard-samples}
\end{figure}

\subsubsection{Datasets} \label{subsubsec:dataset}
We conduct experiments on two datasets. The first one is the
widely used Buffy dataset~\cite{fer08} consisting of 748 annotated video frames
from Buffy TV show. This dataset is proposed as a standard one for HPE task but
not originally for GAC task. We manually annotate the garment attributes for the Buffy
dataset. The second dataset, called ``DL'', contains 1000
daily life photos we collect from websites. Compared with Buffy, the DL dataset possesses
more various garment attribute values. In order to obtain quantitative evaluation results, we
also manually annotate the human parts and garment attributes for the images in
the DL dataset.

Some garment attributes cannot be labeled for the
images in which the person does not wear any garment or some attributes'
visual cues cannot be described. In Figure~\ref{fig:hard-samples}, we illustrate
some of such samples and list the statistical information for the attributes we annotate in
Figure~\ref{fig:cloth-stat}.

\subsubsection{Baselines}
\label{subsubsec:baselines}
We select four state-of-the-art HPE methods as our
baselines: Andriluka et al. (ARS)~\cite{cvpr09}, Sapp et al. (STT)~\cite{eccv10}, Yang and Ramanan (YDR)~\cite{deva11} and Ladicky tl al. (LTZ)~\cite{ladicky}.
As the code of LTZ is not publicly available, we only evaluate our DL dataset by the first three methods.

Although all these algorithms are primarily designed for HPE, they can actually produce results for GAC: as discussed in~\cite{clothliu,clotheccv}, the results of HPE
can be used for part alignment, which enables the extraction of attribute
features. Then for each attribute, we individually train an SVM
multi-class model~\cite{liblinear}. {We use the features described in Table~\ref{tab:cloth-dep} to train each SVM model. We also compare our GAC results with CGG~\cite{clotheccv}~\footnote{This code is not publicly available. We thank the authors Chen et al.~\cite{clotheccv} for providing us the source code for performance comparison.}, which designed a specific pipeline to recognize semantic clothing attributes.}

\subsubsection{Evaluation Metrics}
We evaluate the HPE results with the
standard metric of Probability of Correct Part (PCP)~\cite{bmvc09}.
The GAC results are evaluated by the Garment Attribute Precision (GAP)
criterion, i.e., the classification accuracy for each garment attribute (there
are $5$ attributes in this work).

\subsubsection{Training/Testing} \label{subsubsec:train-test}
For the Buffy dataset,
like~\cite{fer08,fer09,deva11}, we select the images from Episode 3, 4 for
training, and Episode 2, 5 and 6 for testing. For our DL dataset, we select
randomly 300 images for training and use the remaining 700 images for testing.

As we have discussed in Section
\ref{subsubsec:dataset}, some images cannot be annotated with some garment
attributes. For an image without the garment attribute $c_j$, we set all the
features related to $c_j$ to a zero vector in Eq.~\eqref{eq:joint-feature} when
training our structured SVM model and skip evaluation on such attributes.
For a person wearing two (or more) garments, we label each garment's attribute
values. Thus the image has several groups of labels in terms of these garments.
When training the model, all the groups of labels are used to construct
the constraints. When testing a new instance, any attribute
value which the algorithm produces is acceptable if the value belongs to any of the groups.

\subsection{Results}

\begin{figure*}
\centering
\includegraphics[width=0.95\linewidth]{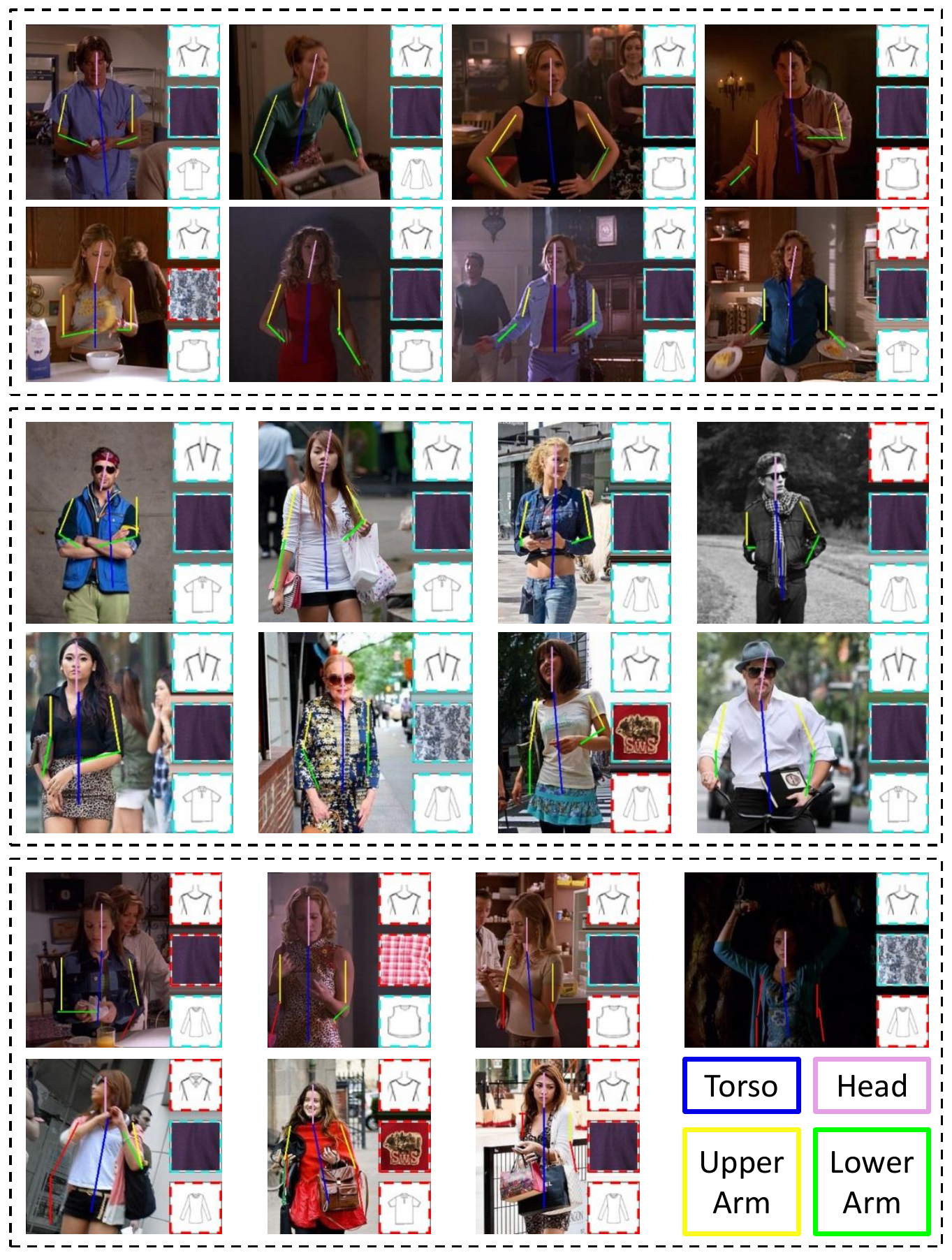}
\caption{\textbf{Examples of our results obtained on the Buffy and DL datasets.} We demonstrate some good results from Buffy and DL in the first and second panels respectively. Some failure cases are showed in the bottom panel. We use the oriented line to denote the pose estimation.
If an HPE result is incorrect, the line is red. We visualize three attributes (neckline, pattern and sleeve) of our GAC results by some icons (see Figure~\ref{fig:attr-def} for the icon definition).
If a GAC result is incorrect, we use a dashed red rectangle to mark it. {Examining the failure cases, we find our algorithm is confused when some human parts are occluded or the human pose is largely variational. Attributes are misclassified when the corresponding parts are mis-detected, or occluded by some objects.}
}\label{fig:result}
\end{figure*}

Figure~\ref{fig:result} shows some exemplar results produced by our
approach. In the following, we shall analyze our approach and compare it with
the baselines.

\begin{figure}
\centering
\subfloat[Buffy]
{
\includegraphics[width=0.6\linewidth]{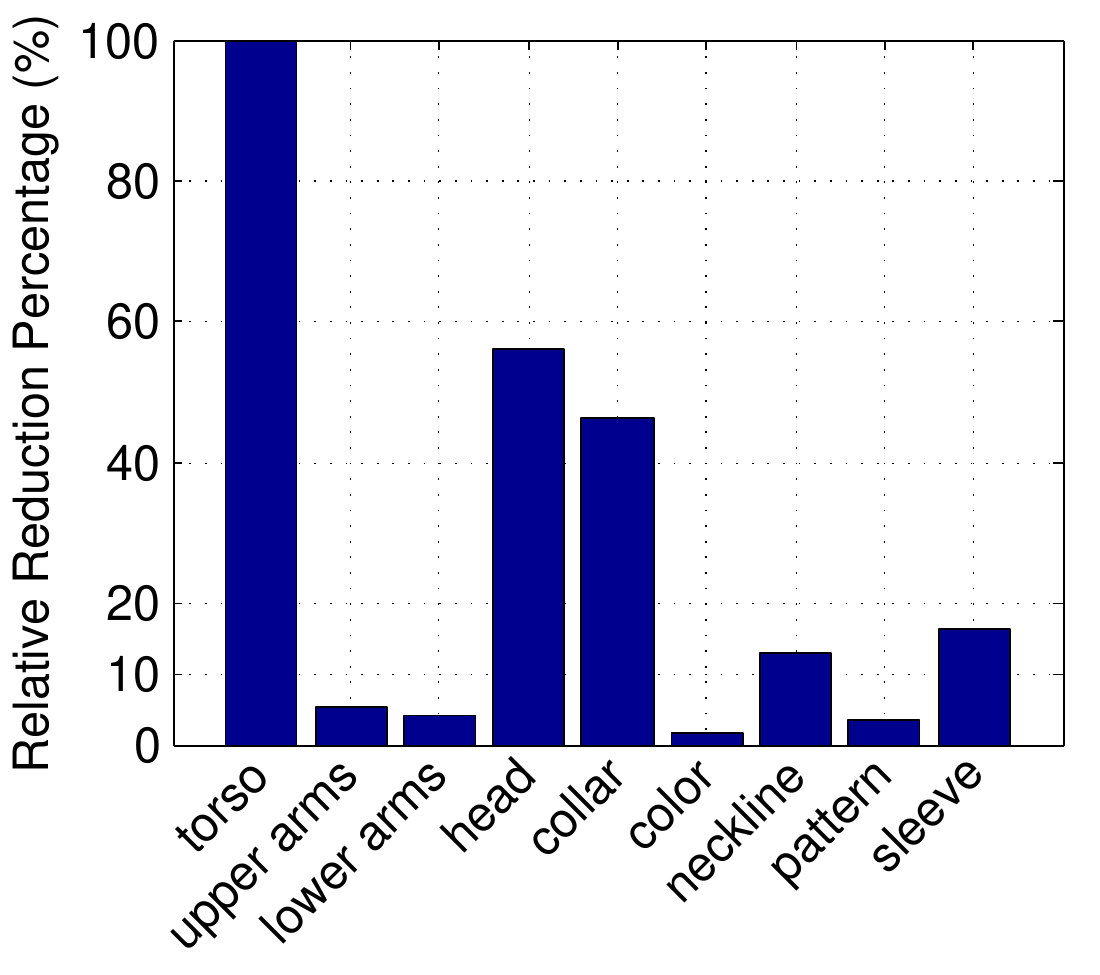}
}

\subfloat[DL]
{
\includegraphics[width=0.6\linewidth]{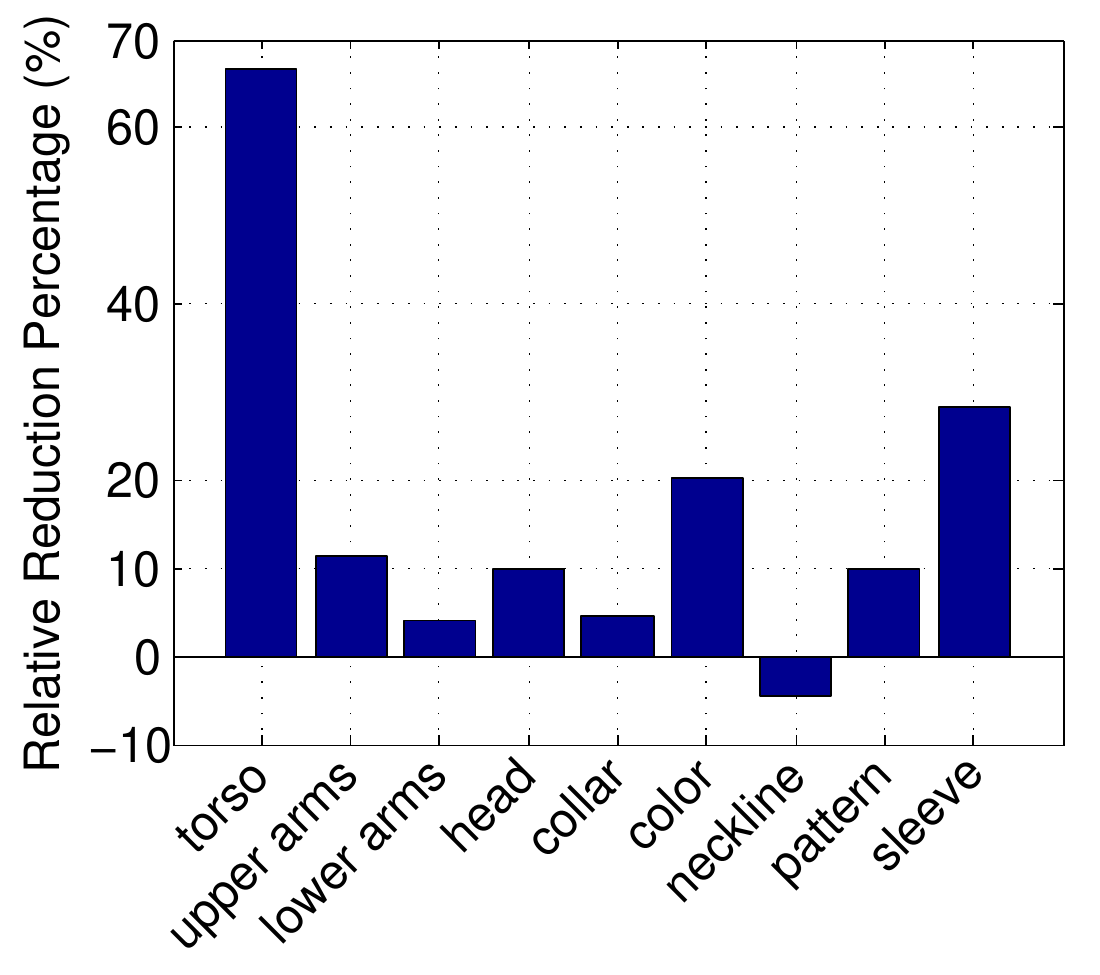}
}
\caption{\textbf{Demonstrating the advantages of combining HPE and GAC together.} X-axis: human part and garment attribute. Y-axis: error reduction rate. We compare our joint learning approach with its separated versions on the Buffy and DL dataset, which deal with HPE and GAC individually. The joint approach improves \textbf{all} of the human parts and a majority of garment attributes. {This is because there exists a strong correlation between human parts and garment attributes, and our algorithm captures their inter-dependency that improves both simultaneously.}}
\label{fig:sep-vs-joint}
\end{figure}

\subsubsection{Examining the Advantages of Joint Learning}
\label{subsubsec:compare-iter}
To show the advantages of combining HPE and GAC together, we compare our joint
learning approach with its two separated versions: one is an HPE algorithm
created by removing the garment-related features in Eq.~\eqref{eq:joint-feature};
the other is a GAC algorithm
created by ignoring all part-related features. Figure~\ref{fig:sep-vs-joint} shows the
comparison results, which demonstrate the significant advantages of the joint
learning over the separated schemes.

Note that the basic appearance constraints of human parts (as \cite{bmvc09} considered)
have been modeled in
the pose-specific features (see Section \ref{subsec:joint-feature}).
Thus, the improvement on HPE is attributed to the integration of garment attribute
evidence modeled in Eq.~\eqref{eq:cross-feature}, i.e., joint inference.
The improvement on GAC can be examined in the same way.

\begin{figure}
\centering
\subfloat[Buffy]
{
\includegraphics[width=0.6\linewidth]{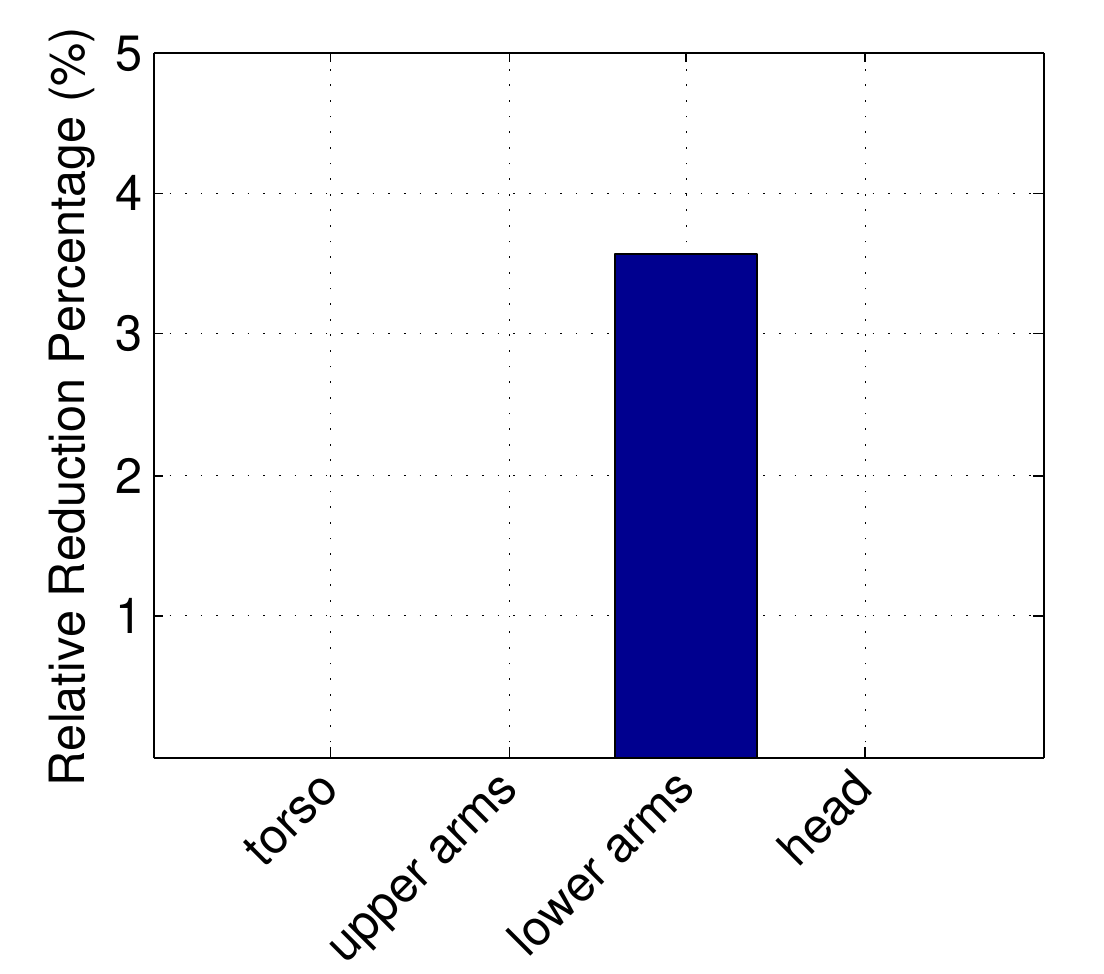}
}

\subfloat[DL]
{
\includegraphics[width=0.6\linewidth]{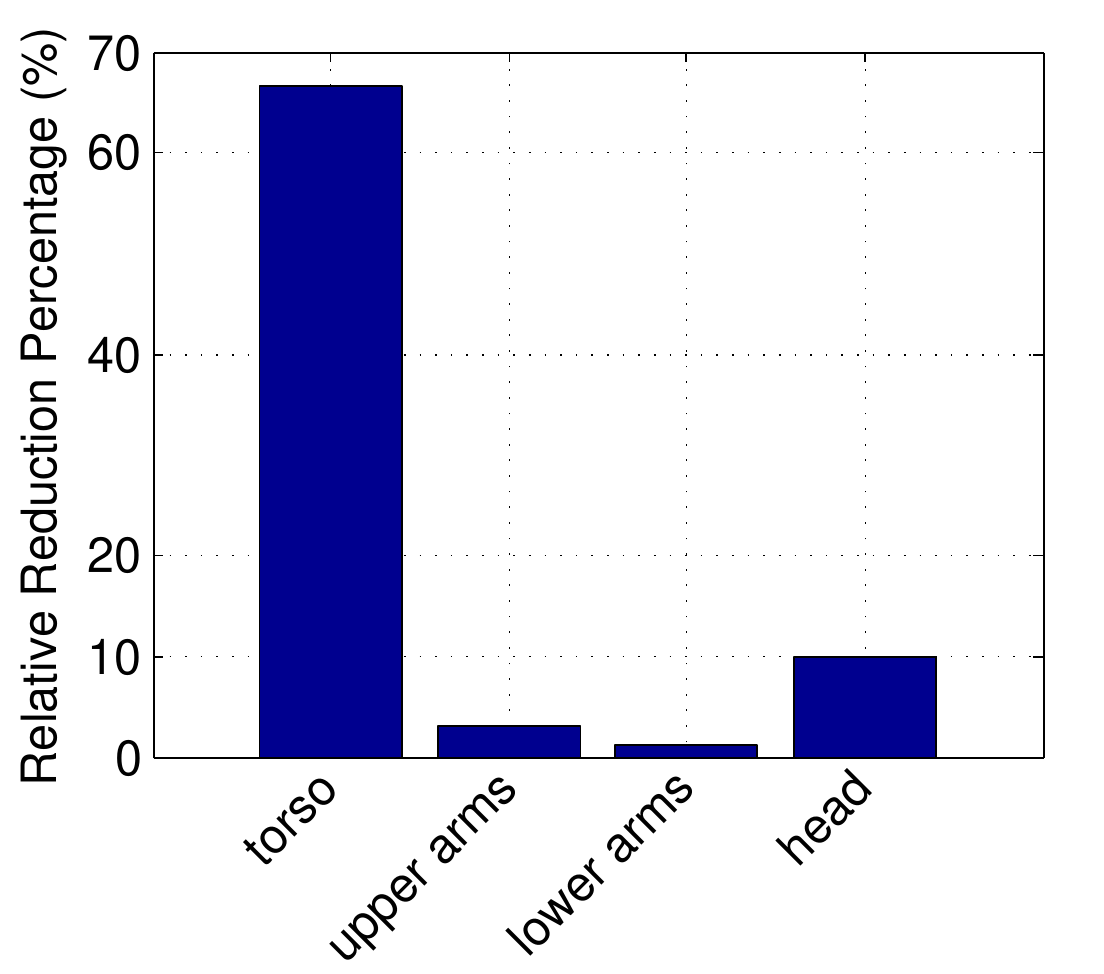}
}
\caption{\textbf{Demonstrating the effectiveness of strong edge evidence.} X-axis: human part. Y-axis: error reduction rate. We demonstrate the error reduction rate for employing the strong edge evidence on both Buffy and DL dataset. {The strong edge evidence improves the detection rate because it captures the contextual information for a human part, which is complementary to other features.}}
\label{fig:noedge-vs-edge}
\end{figure}

\subsubsection{Examining the Effectiveness of Strong Edge}
We demonstrate the effectiveness of the strong edge evidence in this section. By setting the $\alpha$ in Eq.~\eqref{eq:F(x,y;w)} with value zero, our inference algorithm~\ref{alg:infer-for-pose} produces the results without strong edge evidence. Then we use 3-fold cross validation to tune the parameters $\alpha$ and $\beta$ in~Eq.~\eqref{eq:E(x,p)}. The error reduction rate for employing the strong edge evidence on two datasets is demonstrated in Figure~\ref{fig:noedge-vs-edge}. For the Buffy dataset, by using the strong edge evidence, the lower arm accuracy is refined and for the DL dataset, \emph{all} of the human parts are predicted more precisely.

\subsubsection{Comparisons with State-of-the-art Algorithms} \label{subsubsec:compare-state}
\begin{table*}
\caption{\textbf{Comparison with state-of-the-art algorithms on the Buffy dataset.}}
\begin{center}
\begin{tabular}{|l|c|c|c|c||c|}
\hline
Method             & Torso           & U. arms          & L. arms            & Head            & Total \\
\hline\hline
ARS~\cite{cvpr09}  & 90.7            & 79.3              & 41.2              & 95.5          & 73.5\\
\hline
STT~\cite{eccv10}  & \textbf{100}   & 95.3              & 63.0              & 96.2            & 85.5\\
\hline
YDR~\cite{deva11}  & \textbf{100}   & 96.6              & 70.9              & 99.6           & 89.1\\
\hline
LTZ~\cite{ladicky} & \textbf{100}   & \textbf{97.5}     &  75.4             & \textbf{100}   & 90.9\\
\hline
{CGG~\cite{clotheccv}}& --            & --                & --                & --             & --\\
\hline
Our Approach & \textbf{100}   & 96.4              & \textbf{78.4}    & 98.9    & \textbf{91.4}\\
\hline
\end{tabular}
\hspace{6pt}
\begin{tabular}{|c|c|c|c|c||c|}
\hline
Collar           & Color           & Neckline       & Pattern       & Sleeve       & Total\\
\hline\hline
70.3             & 71.7             & 68.7            & 80.9        & 46.6          & 67.7\\
\hline
77.8             & 71.2             & 73.4           & 80.1         & 49.2          & 70.3\\
\hline
 82.8            &  70.8           &  68.3          & 80.9          & 51.5        & 70.9\\
\hline
--                & --             & --              & --           &--            &--\\
\hline
\textbf{89.1}    & 58.4            & 69.4            & 80.9         & 46.2         & 68.8\\
\hline
{88.3}    &\textbf{73.1}    &\textbf{76.1}   &\textbf{81.6}  &\textbf{61.7} &\textbf{76.2}\\
\hline
\end{tabular}
\end{center}

\label{tab:pcp-bf}
\end{table*}

\begin{table*}
\caption{\textbf{Comparison with state-of-the-art algorithms on the DL dataset.}}
\begin{center}
\begin{tabular}{|l|c|c|c|c||c|}
\hline
Method             & Torso           & U. arms          & L. arms     & Head        & Total \\
\hline\hline
ARS~\cite{cvpr09}  & 89.4            & 80.3            & 60.6         & 85.0        & 76.0\\
\hline
STT~\cite{eccv10}  & \textbf{99.9}   & 91.1            & 69.2         & 97.0        & 86.2\\
\hline
YDR~\cite{deva11}  & \textbf{99.9}   & 96.0             & 82.2        & 99.0        & 92.5\\
\hline
{CGG~\cite{clotheccv}}& --            & --                & --                & --             & --\\
\hline
Our Approach  & \textbf{99.9}   & \textbf{96.9}    & \textbf{83.4} & \textbf{99.1}  & \textbf{93.3}\\
\hline
\end{tabular}
\hspace{6pt}
\begin{tabular}{|c|c|c|c|c||c|}
\hline
Collar           & Color           & Neckline       & Pattern         & Sleeve       & Total\\
\hline\hline
70.0             & 55.8             & 50.9          & 77.9            & 60.7         & 63.1\\
\hline
55.5             & 58.1             & 35.9          & 77.7            & 61.9         & 57.8\\
\hline
 75.0            &  58.6           & \textbf{60.0}  & 77.7            & 64.1          & 67.1\\
 \hline
78.1             & \textbf{69.5}   & 59.2           & 78.7           & \textbf{68.4}   & \textbf{70.8}\\
\hline
\textbf{78.5}    &{67.1}          &\textbf{60.0}   & \textbf{79.9}   &{68.2}          &{70.7}\\
\hline
\end{tabular}
\end{center}
\label{tab:pcp-dl}
\end{table*}
In this section, we compare our joint approach (with strong edge evidence) with the state-of-the-art algorithm.
Note that the results of GAC produced by HPE algorithms have been explained
in Section~\ref{subsubsec:baselines}.
Figure~\ref{fig:deva-compare-example} gives the exemplar comparison of HPE and GAC
results from YDR~\cite{deva11} and our joint approach.
On the Buffy dataset,
Table~\ref{tab:pcp-bf} shows that our approach consistently outperforms
YDR~\cite{deva11} which is a recently established algorithm and provides the
candidates for our approach. We also compare our approach with LTZ~\cite{ladicky}
which combines pose estimation and segmentation for computation. We improve the
lower arms performance and achieve the highest overall accuracy.
Table~\ref{tab:pcp-dl} shows the comparison results on the DL dataset. It can be
seen that our approach outperforms all the competing baselines on the task of HPE.

To examine the effectiveness of our approach on GAC, we also compare it with CGG~\cite{clotheccv}, which is a real GAC method. On the Buffy dataset, there is a significant improvement of our approach on the attributes of ``color'' and ``sleeve'', and on the DL dataset, we also reach a competitive performance. The reason of the surprising gain on the Buffy dataset is that human pose of Buffy is more variational than DL. And our model can capture the inter-dependency between human part(s) and garment attributes. However, the pipeline of CGG is step by step, like the work of~\cite{cloth12}. Thus it is depressed if the human pose is largely unconstrained.

One may notice that in Table~\ref{tab:pcp-bf} and Table~\ref{tab:pcp-dl}, compared with the baselines except CGG~\cite{clotheccv}, the GAC results of our approach are significantly improved when we gain less improvement for HPE on average (see YDR~\cite{deva11} for example). The reasons are twofold. First, the training procedure of our approach is different from that of theirs. Our model is trained in a unified manner, which allows us to integrate more useful features. That is, during the training procedure, our model captures the dependency between human parts and garment attributes (i.e. cross-task features), as well as that between different garment attributes (i.e. garment-specific features). As a result, we finally have a global optimal model for HPE and GAC. However, the competing baselines are trained in a separate manner. That is, given the HPE, each garment attribute is trained individually. Therefore, neither the inter-dependency between human parts and garment attributes, nor that between different garment attributes can be utilized, resulting in a marginal model for multiclass SVM. Second, we search for the optimal prediction for GAC by iteratively updating HPE and GAC, reaching a (local) optimal state of HPE and GAC~\footnote{Empirically, the local optima are good enough as we have demostrated in our experiments.}. However, the baselines can only make a prediction for GAC by the given HPE result.

\section{Conclusions and Future Work}\label{sec:con}
Based on the observation that there exist correlations between human parts and
garment attributes, we propose to integrate HPE and GAC into a unified
procedure and handle both tasks simultaneously. We show that such integration
can be seamlessly achieved by using the framework of structured SVM. First,
due to the joint feature representation, it is convenient to involve various
visual cues such as pose-specific features, garment-specific features and
cross-task features. Second, the structured nature of the output space of
structured SVM provides us a straightforward way to jointly infer the solutions
for several problems (e.g., HPE and GAC). Benefiting from these superiorities,
our approach achieves state-of-the-art performance in both HPE and GAC problems,
as demonstrated in the experiments. Obviously, the boosted performance
can benefit quite many multimedia applications, e.g., online clothing retrieval,
clothing recommendation, and we are planning to extend our proposed approach
for these applications in our future work.

\ifCLASSOPTIONcaptionsoff
  \newpage
\fi



\bibliographystyle{IEEEtran}
\bibliography{pe_gac}
%



%






\end{document}